\documentclass[pdflatex,sn-mathphys-num]{sn-jnl}

\usepackage{graphicx}
\usepackage{placeins}
\usepackage{multirow}
\usepackage{amsmath,amssymb,amsfonts,amsthm}
\usepackage{mathrsfs}
\usepackage[title]{appendix}
\usepackage{textcomp}
\usepackage{manyfoot}
\usepackage{booktabs}
\usepackage{algpseudocode}
\usepackage{listings}
\usepackage{array}
\usepackage{soul}

\usepackage[caption=false,font=normalsize,labelfont=sf,textfont=sf]{subfig}
\usepackage{stfloats}
\usepackage{url}
\usepackage{verbatim}
\usepackage[normalem]{ulem}
\usepackage{comment}
\usepackage[table,xcdraw]{xcolor}
\usepackage{threeparttable}
\usepackage{nicematrix}
\usepackage[ruled,vlined]{algorithm2e}
\usepackage{hyperref}

\newcolumntype{G}{>{\columncolor{gray!20}}c} 

\newtheorem{theorem}{Theorem}

\newtheorem{remark}{Remark}

\theoremstyle{thmstyleone}

\theoremstyle{thmstyletwo}

\theoremstyle{thmstylethree}

\begin{document}

\title[Coordinated Target Adaptation]{Biologically Inspired Coordination of Parameter Selection and Update Strength for Language Model Adaptation}

\author[1,3]{\fnm{Xiao} \sur{Zhang}}

\author*[1,3]{\fnm{Tianyu} \sur{Hu}}\email{tianyu@ustb.edu.cn}

\author[1]{\fnm{Juntao} \sur{Lyu}}

\author[2]{\fnm{Qianchuan} \sur{Zhao}}

\author*[1]{\fnm{Huimin} \sur{Ma}}\email{mhmpub@ustb.edu.cn}

\affil[1]{\orgdiv{School of Computer and Communication Engineering}, \orgname{University of Science and Technology Beijing}, \orgaddress{\city{Beijing}, \postcode{100083}, \country{China}}}

\affil[2]{\orgdiv{Department of Automation}, \orgname{Tsinghua University}, \orgaddress{\city{Beijing}, \postcode{100084}, \country{China}}}

\affil[3]{These authors contributed equally: Xiao Zhang, Tianyu Hu}

\abstract{Adapting a source-trained language model to a target task requires deciding which parameters to update and how strongly when learning signals are uncertain. We introduce Synapse Consolidation (SyCo), a biologically inspired framework that coordinates these two controls. Rac1-associated forgetting and memory-trace redistribution motivate retaining selected source-derived low-rank components while resetting others for target adaptation. MAPK-associated signal processing motivates combining multiple prediction-based indicators over time to regulate update strength. SyCo combines these controls with a structured consistency objective and a small, fixed buffer of labelled source examples. Target labels are not used during adaptation. We evaluate the framework in Multi-source Open-set Adaptation (MOA), using source-initialized target learners for tasks absent from source training and for perturbed held-out inputs from source tasks. Across 18 natural-language-processing datasets, SyCo reaches 78.01\% average accuracy on unseen-task adaptation and 85.37\% under unseen-data shifts, outperforming the strongest evaluated hybrid baselines. Ablations support complementary contributions from selective plasticity, evidence-guided update control and structured consistency. These results show how coordinated control of update location and strength can improve target adaptation under uncertain learning signals.}


\maketitle

\section{Introduction}

Large language models (LLMs) can transfer representations and reasoning patterns across tasks \cite{naveed2025llm,ye2024cross,budnikov2025generalization}. Multi-task learning offers one way to build on this capability by training a model on several source tasks, providing a source-trained starting point for target adaptation \cite{mtl,talora}. However, target inputs may differ from source data, and the target task itself may be absent from source training \cite{kejriwal2024nmi}. Adapting to these differences requires modifying the model while retaining source-derived structure that remains useful for the target task. When target labels are unavailable, adaptation often relies on signals derived from the model's own predictions, which can be incorrect. Such signals may therefore drive changes that impair rather than improve target performance. This raises a central question: how should a learner jointly control which parameters are updated and the strength of those updates when adapting under uncertain learning signals?

Test-Time Adaptation (TTA) addresses adaptation at deployment by updating models on unlabeled or self-supervised test data \cite{tta,tent,clust3,foa,poem,t2ard}. Existing TTA methods have made substantial progress on distribution shift, including entropy-based adaptation, information-invariant training, forward-only adaptation and protected online updates \cite{tent,clust3,foa,poem}. However, when adaptation extends to previously unseen tasks, a learner must determine how to reuse source-trained representations while accommodating new task requirements. This creates two related challenges. First, unrestricted updates can alter source-derived structure that remains useful for the target task, whereas overly restrictive protection can hinder the changes needed for adaptation. Second, confident or consistent predictions can still be incorrect, creating a risk of reinforcing errors during adaptation without target labels. Restricting which parameters can change does not by itself resolve this uncertainty, and assessing prediction reliability does not determine which source-derived components should be retained. Effective adaptation therefore calls for coordinated control over where updates occur and how strongly the available evidence drives them.

We study this problem within a source-to-target adaptation episode. A target learner is initialized from a relevant stored source learner and adapted to its target inputs. The source learner remains available separately. Our objective is to improve target adaptation by reusing source knowledge and regulating parameter changes according to the available evidence. Selective retention serves this objective rather than an independent requirement to preserve all source-task capabilities. Unlike continual learning, our evaluation does not assess cumulative acquisition and retention across successive tasks. We therefore evaluate whether coordinating selective parameter updates with evidence-guided control of update strength improves target-task performance.

Biological principles of plasticity have inspired computational approaches to adaptive learning, including continual learning \cite{zhongyi-nmi} . Here, we draw on complementary biological processes to motivate coordinated control over parameter selection and update strength during target adaptation. Genetic dissection of memory interference in Drosophila identifies distinguishable contributions from Rac1 and the CSW/Raf/MAPK pathway to interference between learning experiences \cite{zhongyi-elife}. Rac1 regulates active forgetting \cite{rac1-active-foregeting1}, and recent work links natural memory decay to a Rac1-dependent redistribution of memory traces between synaptic ensembles before their later elimination \cite{rac1-space}. Studies of activity-dependent synapse elimination provide a related motivation for selective retention and weakening \cite{activity2,activity1}. These biological observations motivate selectively retaining some components while allowing others to change. We translate this structural principle into a plasticity confiner: selected low-rank components of a retrieved source learner are retained, while the coefficients of other components are reset to zero and those components are opened for target adaptation. Our experiments test whether the resulting component-selection rule improves target adaptation.
\begin{figure} 
        \centering
        \includegraphics[width=1.0\textwidth]{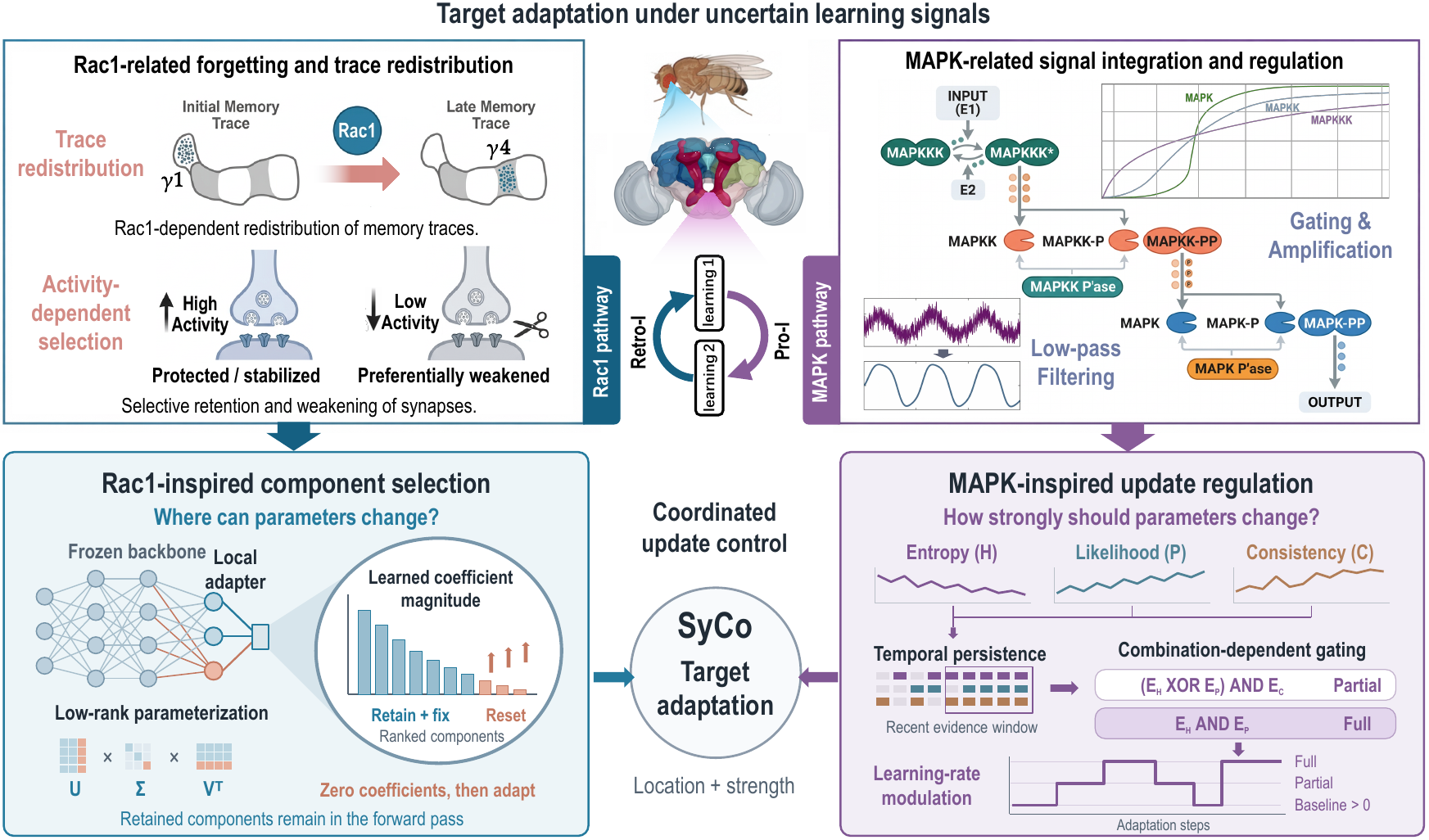}
        \caption{Biological motivation for coordinating update location and strength during target adaptation. Rac1-related forgetting and memory-trace redistribution motivate selectively retaining some model components while allowing others to change. In the computational design, selected low-rank components of a retrieved source learner are retained while the coefficients of other components are reset to zero and opened for target adaptation. MAPK-related signal integration, coincidence detection, filtering and amplification motivate combining multiple prediction-based indicators over time to regulate update strength. The two mechanisms address which parameters may change and how strongly incoming evidence drives those changes. Pro-I and Retro-I describe the biological interference phenomena motivating the design. The diagrams illustrate functional abstractions rather than a one-to-one correspondence between molecular pathways and parameter dynamics.} \label{fig:biological_inspiration}
\end{figure}
MAPK-related processes motivate control over the temporal integration and strength of learning signals. Learning-evoked Raf/MAPK activity can protect labile memory against disruption \cite{rac-mapk}. Studies of MAPK/ERK signalling also describe signal integration and coincidence detection, with concurrent activation of upstream inputs producing sustained ERK responses \cite{mapk-integrate}. These properties, together with noise filtering, threshold-like responses and amplification \cite{mapk-filter,mapk-gate2,mapk-amplify1,mapk-amplify2,mapk-gate1}, motivate combining multiple prediction-based indicators over time to regulate update strength. In SyCo, entropy-, likelihood- and consistency-based evidence is temporally filtered, and its combinations determine the learning-rate modulation. Temporal filtering reduces sensitivity to transient changes, while the activation rule assigns different update strengths to different combinations of evidence. In this computational sense, the controller is metaplastic: it regulates how subsequent plasticity is expressed \cite{meta-plasticity}. The biological observations motivate the separation and coordination of structural selection and signal-dependent regulation (Fig.~\ref{fig:biological_inspiration}). They do not prescribe the specific parameter ranking or logical gating rule used in the algorithm.

\begin{figure}[ht!]
    \centering
    \includegraphics[width=0.95\textwidth]{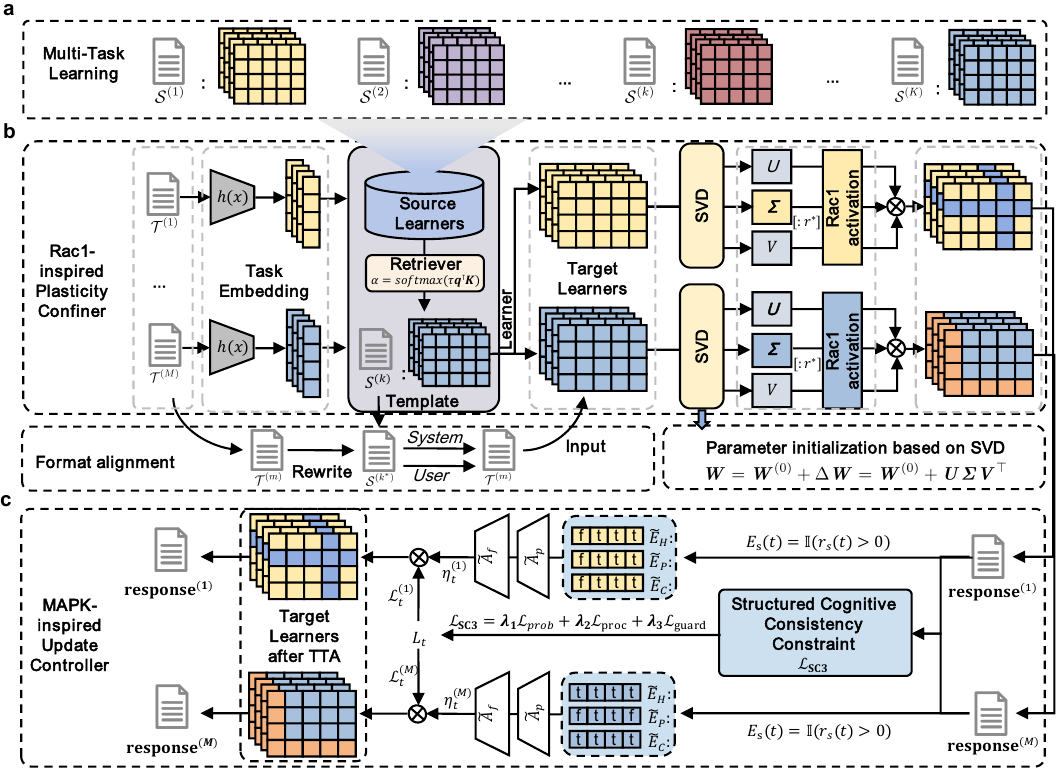}
    \caption{
    \textbf{SyCo integrates transferable cross-task knowledge with biologically inspired adaptive plasticity control.}
    \textbf{a,} In the multi-task learning stage, SyCo trains task-specific source learners from a set of source tasks and stores their task representations, prompts and parameter states as reusable priors. These priors form the memory bank used when a new target task is encountered.
    \textbf{b,} For each target task, the Rac1-inspired plasticity confiner first encodes the task description and retrieves the most relevant source learner and template. The retrieved template is rewritten into a target-compatible system--user format to construct the input prompt, while the matched source learner initializes the target learner. The initialized parameters are then organized into low-rank adaptation subspaces using the low-rank parameterization with learnable coefficients described in Methods. Rac1-inspired activation opens the selected tail components for updating, thereby localizing plasticity to a subset of low-rank components. The coloured regions after Rac1 activation indicate the activated subspaces that receive task-specific updates.
    \textbf{c,} During online adaptation, target learners generate responses that are assessed by structured cognitive signals. The structured cognitive consistency constraint, $\mathcal{L}_{\text{SC3}}$, generates adaptation signals through problem understanding, process reasoning, and a labelled source-data guardrail. The MAPK-inspired regulator then uses entropy-, likelihood-, and consistency-based evidence to trigger partial or full activation events, thereby scaling the strength of the update signals. After the controlled updates, the adapted target learners produce the final task-specific responses.}

    \label{fig:framework}
\end{figure}

Integrating these mechanisms, we develop Synapse Consolidation (SyCo) for target adaptation (Fig.~\ref{fig:framework}). Source retrieval and prompt alignment provide the target learner with a relevant initialization and input format. The Rac1-inspired plasticity confiner determines which low-rank components may change, and the MAPK-inspired controller regulates their update strength. A Structured Cognitive Consistency Constraint ($\mathcal{L}_{\mathrm{SC3}}$) supplies the adaptation objective, drawing on semantic invariance under input variation \cite{gao2021emnlp}, consistency-based learning signals \cite{wang2022iclr}, and a guardrail using labelled source examples. Target labels are not used during adaptation. The combined design addresses two different sources of difficulty: useful source structure may be altered by unrestricted updates, and even a restricted update can be driven by misleading evidence. The experiments examine the contributions of both controls and of the learning objective within the same target-adaptation framework.

\begin{figure}[htbp]
        \centering
        \includegraphics[width=0.95\textwidth]{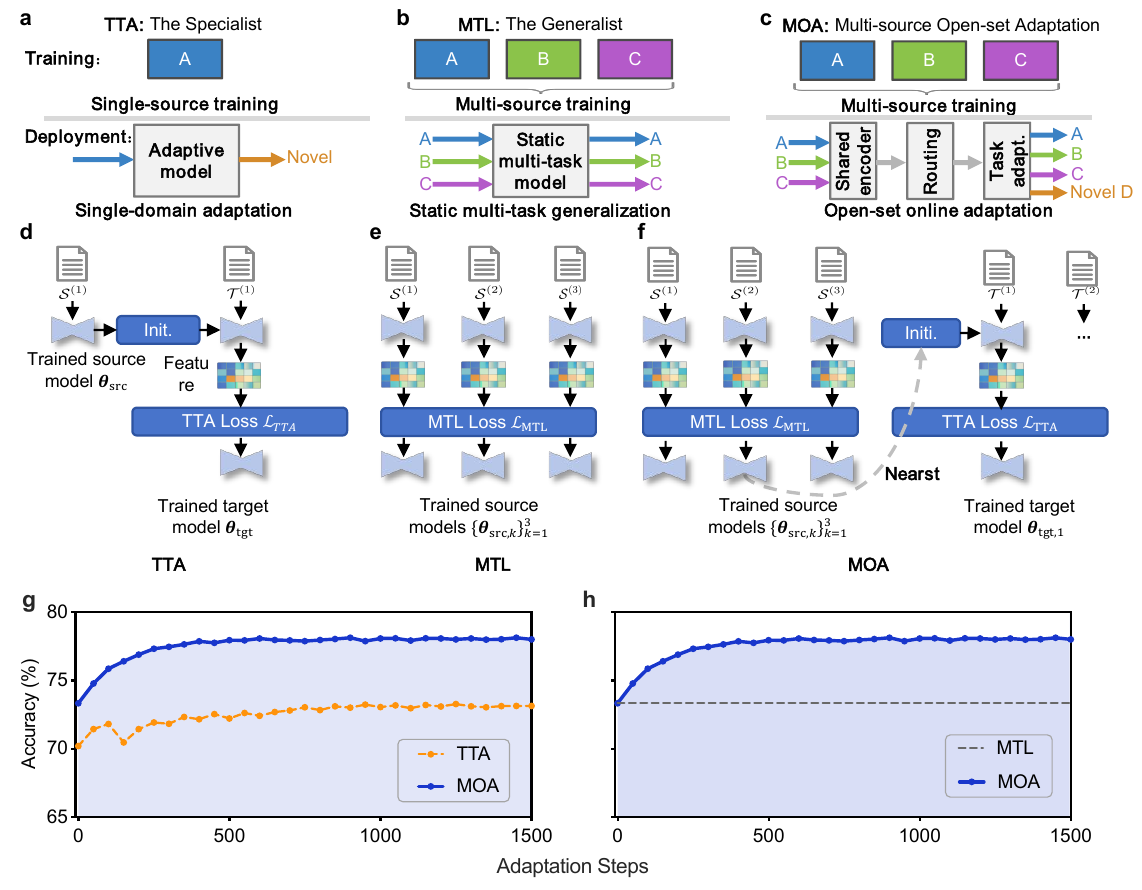}
        \caption{Schematic framework and empirical comparison of different paradigms under non-stationary deployment.
        Here, TTA, MTL, and MOA denote Test-Time Adaptation, Multi-Task Learning, and Multi-source Open-set Adaptation, respectively. Note that ``adapt.'' serves as an abbreviation for adaptation.
        \textbf{a}--\textbf{c}, Conceptual comparison of model deployment strategies, illustrating single-domain TTA as a specialist (\textbf{a}), static MTL generalization as a generalist (\textbf{b}), and the proposed open-set MOA paradigm (\textbf{c}).
        \textbf{d}, Technical pipeline of TTA (exemplified by $\text{T}^2\text{ARD}$), where a single-source model $\boldsymbol{\theta}_{\text{src}}$ is adapted to an unlabeled target stream $\mathcal{T}^{(1)}$ using a self-supervised loss $\mathcal{L}_{\text{TTA}}$.
        \textbf{e}, Technical pipeline of MTL (exemplified by TA-LoRA), where a static model is jointly trained on multiple source tasks ($\mathcal{S}^{(1)}$, $\mathcal{S}^{(2)}$, $\mathcal{S}^{(3)}$) via $\mathcal{L}_{\text{MTL}}$.
        \textbf{f}, Technical pipeline of MOA, which integrates a shared encoder, routing modules, and task-specific adaptation during training and open-set deployment.
        \textbf{g}, Accuracy comparison between TTA ($\text{T}^2\text{ARD}$) and the proposed MOA (SyCo) over 1,500 adaptation steps on a mixed target stream ($\mathcal{T}^{(1)}$, $\mathcal{T}^{(2)}$).
        \textbf{h}, Accuracy comparison showing the performance trajectories of static MTL (TA-LoRA) and MOA  (SyCo) under non-stationary task shifts. These comparisons evaluate target adaptation from source-derived initialization.}
        \label{fig:methods_comparison}
\end{figure}

We evaluate SyCo in a Multi-source Open-set Adaptation (MOA) setting (Fig.~\ref{fig:methods_comparison}). Models trained on eight source tasks are adapted to ten target tasks absent from source training or to perturbed held-out inputs from the source tasks. Here, open-set refers to target tasks outside the source-training task set. Target-task descriptions are available for source retrieval and prompt alignment. Across 18 NLP datasets, SyCo outperforms the evaluated TTA, multi-task-learning and hybrid baselines in both unseen-task and unseen-data settings. Comparisons of masking rules, controller ablations and objective ablations assess how selective parameter updates and evidence-guided regulation contribute to target performance.

\section{Results}
\subsection{MOA Exposes Complementary Limits of TTA and MTL}
MOA combines two deployment challenges: learning transferable structure from multiple labelled source tasks and adapting to unlabeled target streams containing distribution shifts and new tasks. In Fig.~\ref{fig:methods_comparison}, the TTA baseline $\text{T}^2\text{ARD}$ \cite{t2ard} adapts online but starts from a single-source representation, whereas the Multi-Task Learning (MTL) baseline TA-LoRA \cite{talora} provides a stronger initialization but remains static at deployment. Their trajectories illustrate complementary limitations: the TTA baseline lacks a multi-task prior, while static MTL cannot update its parameters in response to target-task or distribution changes. SyCo combines source-informed initialization with regulated test-time updates. On the mixed target stream, it starts from a stronger initialization than the TTA baseline and improves beyond static MTL during adaptation (Fig.~\ref{fig:methods_comparison}g,h). Allowing updates, however, does not by itself ensure stable adaptation. Without the Rac1-inspired confiner, unseen-data performance initially improved but later declined during prolonged adaptation, suggesting that unrestricted updates can undermine early gains (Fig.~\ref{fig:rac1}b). These observations motivate examining how structural protection and update regulation contribute to target adaptation.

\subsection{Synapse Consolidation Framework}
SyCo separates adaptation into three interacting components (Figs. \ref{fig:biological_inspiration} and \ref{fig:framework}). The Rac1-inspired plasticity confiner determines where parameters may change by restricting updates to a selected low-rank subspace. The MAPK-inspired controller determines the strength of each update by integrating reliability evidence over time and gating the effective learning rate. A Structured Cognitive Consistency Constraint ($\mathcal{L}_{\mathrm{SC3}}$) supplies the adaptation objective through problem-level invariance, process-level calibration and a labelled source-data guardrail, without using target labels. This separation is deliberate: selecting which parameters may change and regulating how strongly uncertain target evidence drives those changes address distinct risks during adaptation. The following ablations test whether structural protection, evidence-dependent update control and the structured learning signal provide complementary benefits for source-to-target adaptation.

\subsection{Plasticity Confinement Supports Source Knowledge Reuse}

\begin{figure}[htb]
        \centering
        \includegraphics[width=0.98\textwidth]{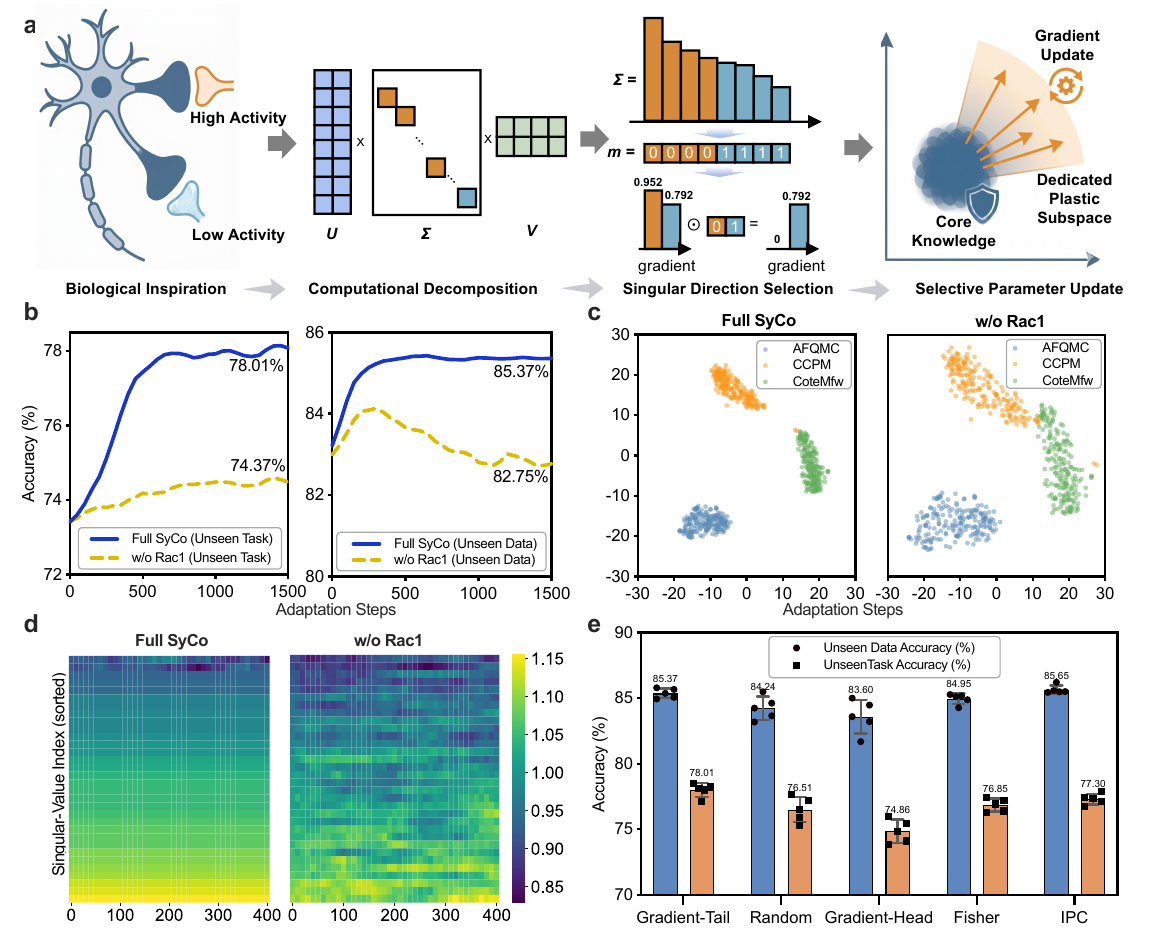}
    
        \caption{
        \textbf{Rac1-inspired plasticity confinement stabilizes test-time adaptation in SyCo.}
        \textbf{a,} Schematic of the Rac1-inspired module. Motivated by activity-dependent biological plasticity, SyCo organizes the adaptable parameters into low-rank components and selects a subset as the dedicated plastic subspace. Gradient updates are then confined to this selected subspace, while the remaining component parameters are held fixed.
        \textbf{b,} Adaptation curves on unseen-task and unseen-data settings. Full SyCo achieves higher and more stable accuracy than the variant without Rac1, reaching 78.01\% and 85.37\%, respectively, compared with 74.37\% and 82.75\% for w/o Rac1.
        \textbf{c,} T-SNE visualization of task representations for AFQMC, CCPM and CoteMfw. Full SyCo maintains compact and well-separated task clusters, while w/o Rac1 leads to more dispersed and partially entangled representations.
        \textbf{d,} Evolution of sorted learned coefficients during adaptation. The vertical index enumerates sorted low-rank components, and colours represent their learned coefficients rather than exact singular values. Full SyCo produces more structured and localized changes along selected low-rank components, whereas removing Rac1 causes broader and noisier perturbations across the whole parameter spectrum.
        \textbf{e,} Ablation of component-selection strategies. Gradient-tail masking outperforms random and gradient-head masking under the same update ratio $\alpha=0.1$, supporting selective parameter updates for target adaptation. The strategy labelled Gradient-tail uses coefficient-magnitude ranking as described in Methods.
        }

        \label{fig:rac1}
\end{figure}

Confining plasticity improved accuracy in both settings: full SyCo reached 78.01\% accuracy on unseen tasks and 85.37\% on unseen-data shifts, compared with 74.37\% and 82.75\%, respectively, without the confiner (Fig. \ref{fig:rac1}b). 

These gains support selective parameter updates for target adaptation, with representation and parameter analyses providing complementary structural evidence. Task representations remained more clearly separated with SyCo in the t-SNE visualization \cite{tsne}, whereas removing the confiner produced greater overlap (Fig. \ref{fig:rac1}c). Tracking the learned coefficients likewise showed more localized changes with full SyCo, whereas removing the confiner led to broader changes across the low-rank components (Fig. \ref{fig:rac1}d). All masking strategies use the same update ratio $\alpha=0.1$, with the remaining directions protected. We compared coefficient-based selection with random and gradient-head masking. Gradient-tail masking achieved 85.37\%/78.01\% for unseen data/tasks, compared with 84.24\%/76.51\% for random masking and 83.60\%/74.86\% for gradient-head masking (Fig. \ref{fig:rac1}e). With the update ratio fixed, these gains support selecting plastic directions rather than merely limiting update capacity. Fisher importance estimates source-loss sensitivity~\cite{kirkpatrick2017nas}, whereas IPC combines smoothed sensitivity and uncertainty~\cite{ling2025ipc}. We compared Fisher and IPC masking at the same update ratio to test whether protecting source-sensitive parameters also benefits target transfer. At the same update ratio, the proposed mask achieved higher mean accuracy on unseen tasks than IPC (78.01\% versus 77.30\%), whereas IPC achieved higher mean accuracy on unseen data (85.65\% versus 85.37\%).

\begin{figure}[!t]
        \centering
        \includegraphics[width=\linewidth,height=0.64\textheight,keepaspectratio]{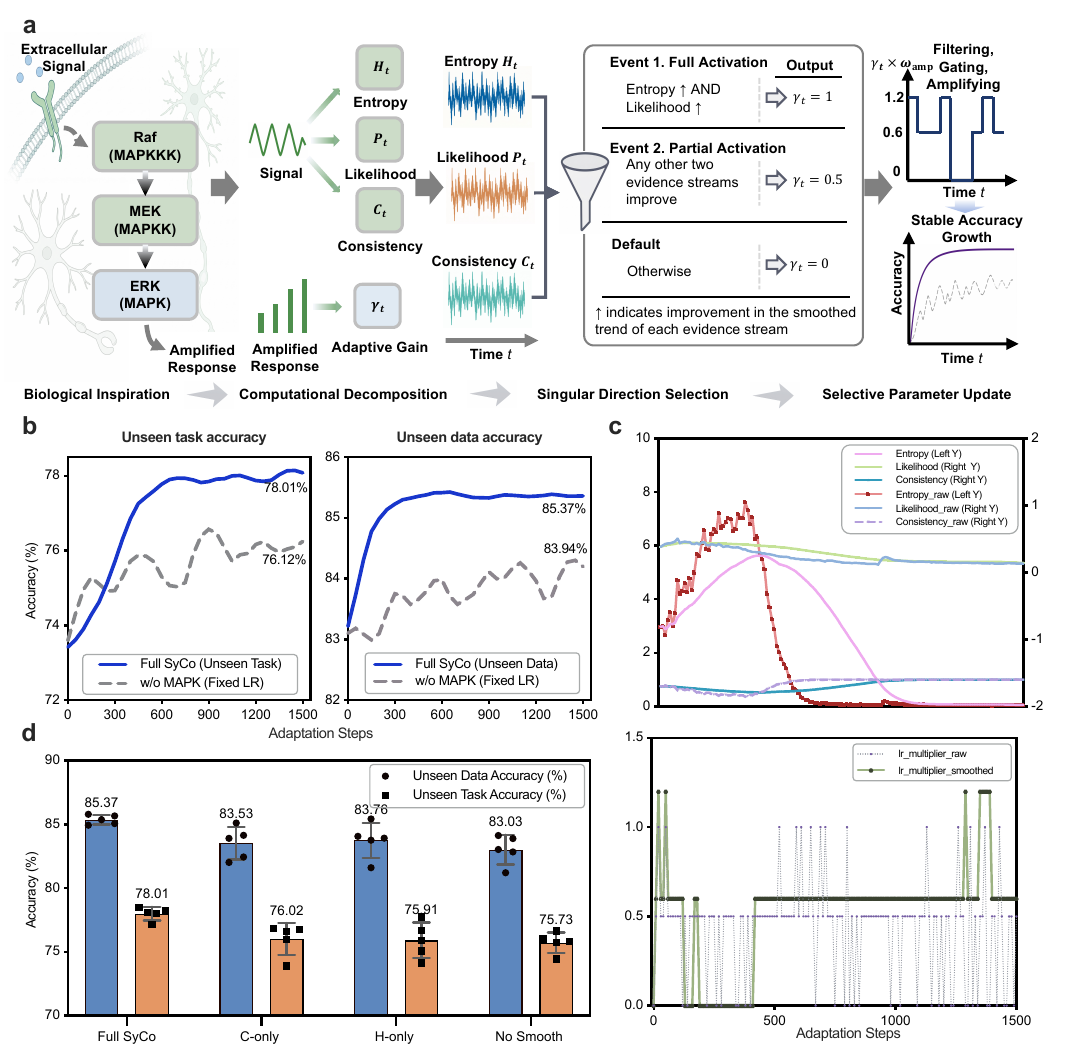}
        \caption{
        \textbf{MAPK-inspired evidence-aware modulation for test-time adaptive updates.}
        \textbf{a,} Workflow of the MAPK-inspired controller. Inspired by the MAPK cascade that filters noisy signals and amplifies reliable responses, SyCo extracts entropy-, likelihood- and consistency-based evidence from target responses. These evidence streams are temporally smoothed by low-pass filtering, passed through an evidence-aware gate to assign full, partial or baseline update states, and converted into an adaptive gain that scales gradient updates according to the filtered evidence.
        \textbf{b,} Adaptation curves in unseen-task and unseen-data settings. Full SyCo improves faster and reaches higher final accuracy than the fixed-learning-rate variant (without MAPK), attaining 78.01\% and 85.37\%, respectively.
        \textbf{c,} Internal controller dynamics. Raw evidence streams are smoothed to estimate stable trends, producing a time-varying learning-rate multiplier across adaptation steps instead of a uniform update strength.
        \textbf{d,} Component ablations of evidence-aware gating. Full SyCo outperforms variants using only consistency evidence, only entropy evidence or no temporal smoothing, indicating that both multi-evidence integration and temporal filtering contribute to target adaptation in these experiments. Black dots and squares denote individual runs for unseen-data and unseen-task accuracy, respectively; error bars denote standard deviation. Equations~\eqref{eq:partial}--\eqref{eq:learning_rate} specify the activation rule and learning rate.
        }

        \label{fig:MAPK}
\end{figure}

\subsection{Evidence-guided Control Stabilizes Test-time Updates}
We next tested whether evidence-guided regulation of update strength improves target adaptation within the selected plastic subspace. Replacing the MAPK-inspired controller with a fixed learning rate reduced unseen-task accuracy from 78.01\% to 76.12\% and unseen-data accuracy from 85.37\% to 83.94\% (Fig. \ref{fig:MAPK}b). The controller combines entropy-, likelihood- and consistency-based evidence, smooths these signals over time and maps their joint trend to baseline, partial or full update states (Figs. \ref{fig:MAPK}a and \ref{fig:MAPK}c). This design reduced rapid fluctuations in the raw evidence streams and produced a time-varying learning-rate multiplier rather than a uniform step size.

Component ablations further showed that the gain did not arise from a single confidence measure. Consistency-only and entropy-only variants reached 83.53\%/76.02\% and 83.76\%/75.91\% on unseen data/tasks, respectively, whereas removing temporal smoothing reduced performance to 83.03\%/75.73\% (Fig. \ref{fig:MAPK}d). The full controller reached 85.37\%/78.01\%. To examine how multi-evidence control regulates update strength under misleading confidence, we evaluated confidently incorrect predictions (confidence $\geq 0.9$). SyCo yielded a mean learning-rate multiplier of $0.24$, compared with $0.57$ for confidence-only control. The lower mean multiplier indicates more conservative updating for the evaluated confidently incorrect predictions.

\begin{figure}[!t]
        \centering
        \includegraphics[width=0.98\textwidth]{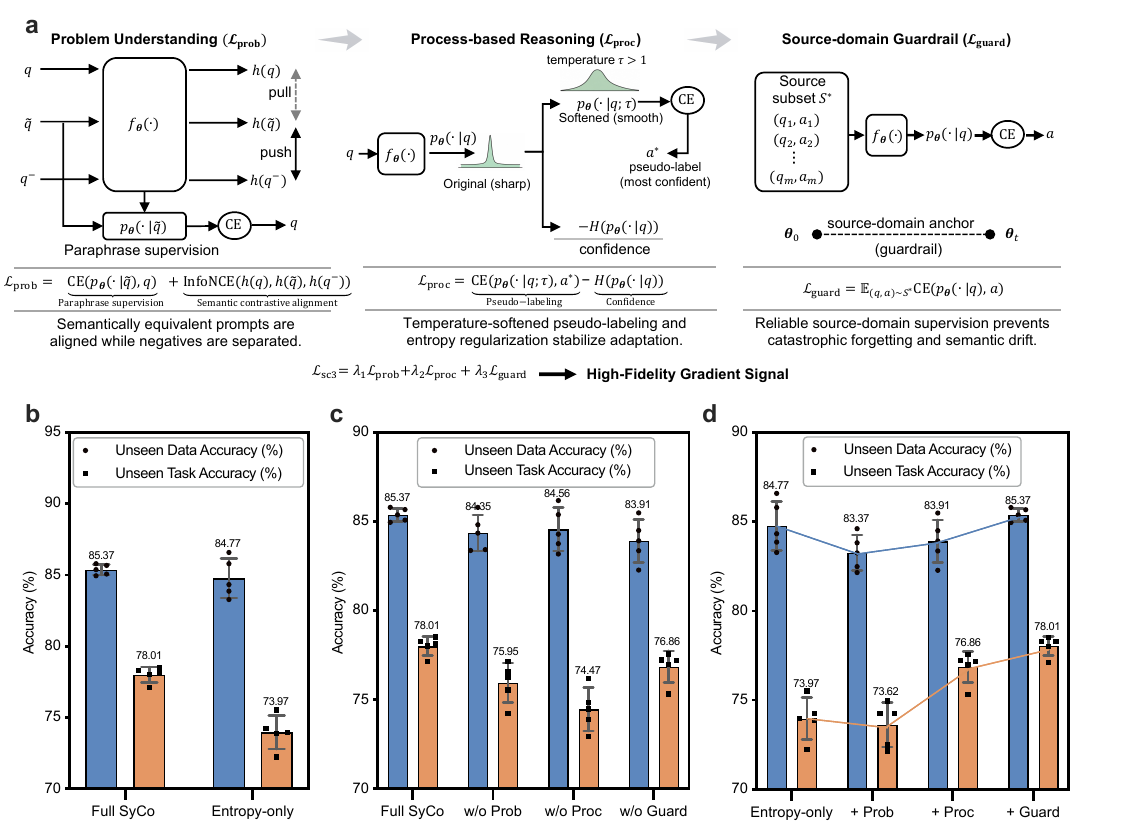}

        \caption{
        \textbf{Structured cognitive consistency constraint stabilizes SyCo's adaptation.}
        \textbf{a,} Anatomy of $\mathcal{L}_{\mathrm{SC3}}$. SyCo constructs three complementary supervision signals: problem-understanding consistency aligns paraphrased prompts and separates semantic negatives, process-based reasoning uses softened pseudo-labels and confidence regularization to stabilize adaptation, and the source-domain guardrail anchors updates to reliable source supervision to limit drift from source-derived knowledge during target adaptation. These terms are combined to provide a structured learning objective for test-time adaptation.
        \textbf{b,} Comparison between Full SyCo and the entropy-only variant. Replacing the full structured consistency design with entropy-only guidance reduces both unseen-data and unseen-task accuracy.
        \textbf{c,} Component ablation of $\mathcal{L}_{\mathrm{SC3}}$. Removing the problem-understanding, process-based reasoning, or source-domain guardrail term degrades performance to different extents, showing that the three constraints provide complementary adaptation signals.
        \textbf{d,} Incremental composition of the constraint. +Prob uses only the problem-understanding term and does not by itself outperform the entropy-only baseline. Adding process-based reasoning and source-domain guardrail components progressively improves performance, with the full constraint performing best. Black dots and squares denote individual runs for unseen-data and unseen-task accuracy, respectively; error bars denote standard deviation.
        }
 \label{fig:LSC3}
\end{figure}

\subsection{A Structured Consistency Objective Improves Target Adaptation}
We then examined the structured objective supplying these updates (Fig.~\ref{fig:LSC3}). Replacing $\mathcal{L}_{\mathrm{SC3}}$ with entropy minimization reduced unseen-data accuracy from 85.37\% to 84.77\% and unseen-task accuracy from 78.01\% to 73.97\% (Fig. \ref{fig:LSC3}b), with the larger drop on unseen tasks indicating that entropy alone is insufficient when the task itself changes.

The three $\mathcal{L}_{\mathrm{SC3}}$ terms contributed differently. Removing problem-level invariance reduced performance to 84.35\%/75.95\%; removing process-level calibration to 84.56\%/74.47\%; and removing the source-domain guardrail to 83.91\%/76.86\% on unseen data/tasks (Fig. \ref{fig:LSC3}c). The progressive composition experiment showed that the problem term alone did not improve the entropy baseline, whereas adding process calibration and then the source guardrail progressively recovered performance to the full 85.37\%/78.01\% result (Fig. \ref{fig:LSC3}d).
\begin{figure}[!t]
        \centering
        \includegraphics[width=0.98\textwidth]{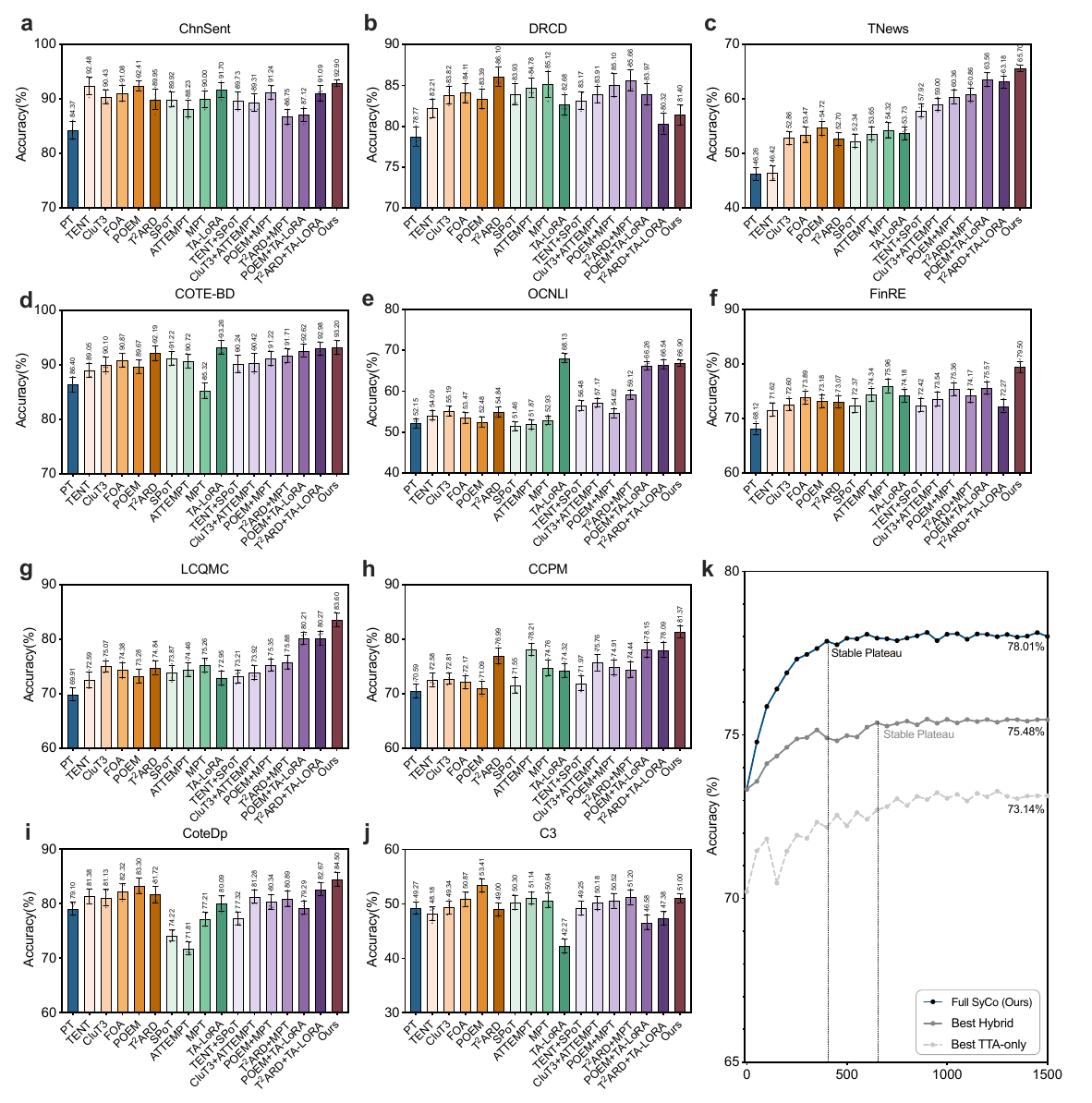}
        \caption{
        \textbf{Unseen-task adaptation performance.}
        \textbf{a--j,} Accuracy comparison on ten unseen tasks, including ChnSent, DRCD, TNews, COTE-BD, OCNLI, FinRE, LCQMC, CCPM, CoteDp, and C3. Baseline methods are grouped into technical families by color codes: orange bars denote the TTA family, green bars represent the MTL family, and purple bars indicate the hybrid family. SyCo achieves the best or near-best accuracy on most unseen-task settings, supporting adaptation to target tasks absent from source training.
        \textbf{k,} Online adaptation dynamics in the unseen-task setting. Full SyCo adapts faster and reaches a higher stable accuracy plateau than the strongest hybrid and TTA-only baselines. Mean benchmark accuracies are 78.01\%, 75.48\% and 73.14\%, respectively. Error bars denote standard deviation.
         }
    \label{fig:UNSEEN_TASK}
\end{figure}
\subsection{Performance Across the MOA Benchmark}
The ablations support complementary contributions from the three components. We next compared the full framework with TTA, MTL and hybrid baselines on unseen tasks and distribution-shifted data.

On ten unseen tasks, SyCo ranked first or second on eight datasets and achieved 78.01\% mean accuracy, exceeding the strongest hybrid baseline, $\text{T}^2\text{ARD}$+TA-LoRA (75.48\%), the TTA-only $\text{T}^2\text{ARD}$ baseline (73.14\%) and the static TA-LoRA initialization (73.33\%) (Fig. \ref{fig:UNSEEN_TASK}a--j). With each batch mixing different target tasks to simulate MOA's open-set setting, SyCo's higher mean accuracy supports its effectiveness on heterogeneous target inputs. Source tasks were excluded from these batches to focus on source-to-target initialization.

\begin{figure}[ht!]
        \centering
        \includegraphics[width=0.98\textwidth]{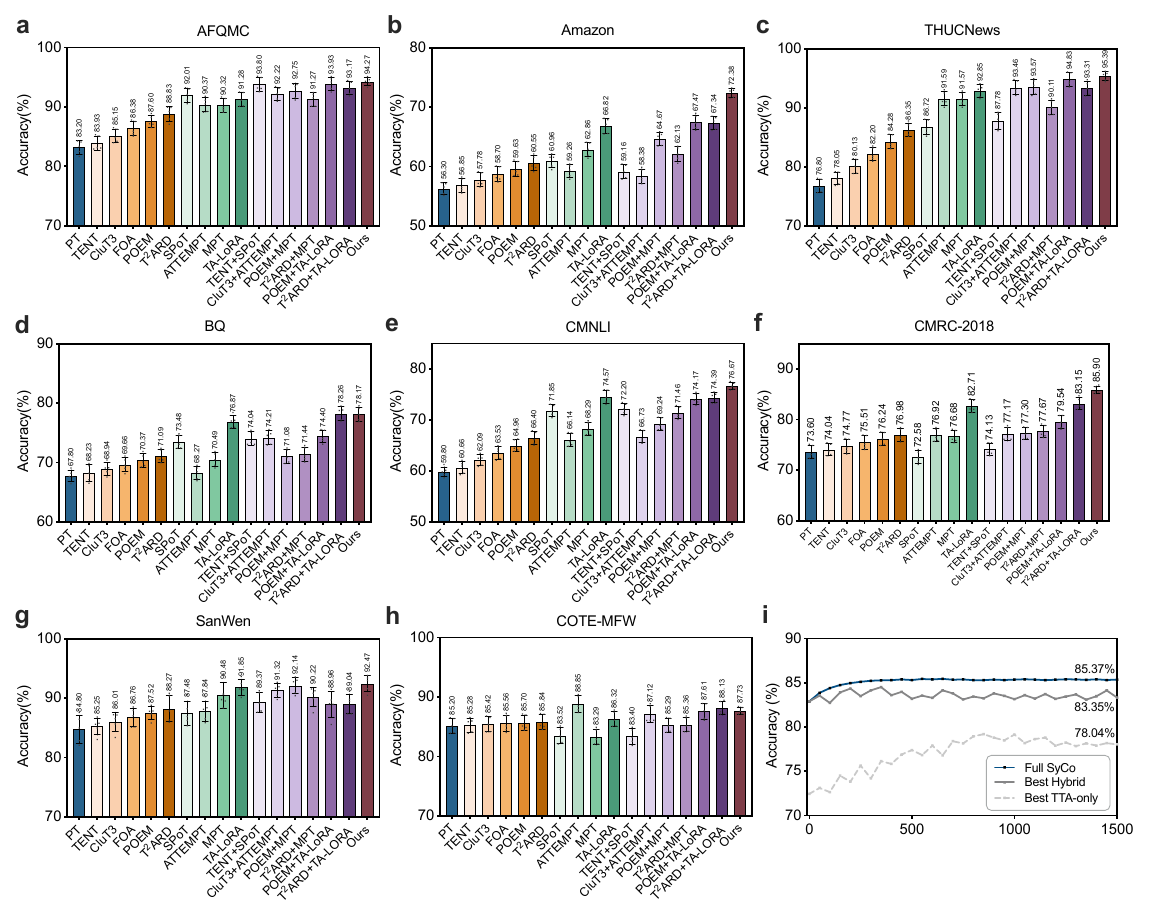}
        \caption{
        \textbf{Unseen-data adaptation performance on source tasks under distribution shifts.} \textbf{a--h}, Accuracy comparison on perturbed held-out inputs from eight source tasks, including AFQMC, Amazon, THUCNews, BQ, CMNLI, CMRC-2018, SanWen, and COTE-MFW. Baseline methods are grouped into technical families by color codes: orange bars denote the TTA family, green bars represent the MTL family, and purple bars indicate the hybrid family. SyCo achieves the best or near-best accuracy on most tasks, supporting adaptation to the evaluated perturbed inputs. \textbf{i}, Online adaptation dynamics in the unseen-data setting. Full SyCo converges faster and reaches a higher, more stable accuracy plateau than the strongest hybrid and TTA-only baselines. Mean benchmark accuracies are 85.37\%, 83.35\% and 78.04\%, respectively. Error bars denote standard deviation.
        }
    \label{fig:UNSEEN_DATA}
\end{figure}

Across eight source-task evaluations using perturbed held-out inputs, SyCo reached 85.37\% mean accuracy, compared with 83.35\% for the strongest hybrid baseline and 78.04\% for the strongest TTA-only baseline (Fig. \ref{fig:UNSEEN_DATA}a--h). Its trajectory also showed a higher and more stable plateau (Fig. \ref{fig:UNSEEN_DATA}i). Because SyCo and these hybrid comparators use the same TA-LoRA initialization method, the gains on unseen tasks and distribution-shifted data support the contribution of SyCo's adaptation modules beyond initialization alone. These gains were reproduced across different pretrained backbones in the supplementary experiments, and the benefits of the multi-source component persisted across model sizes (Supplementary Note~1).

\begin{figure}[htbp]
\centering
\includegraphics[width=\textwidth]{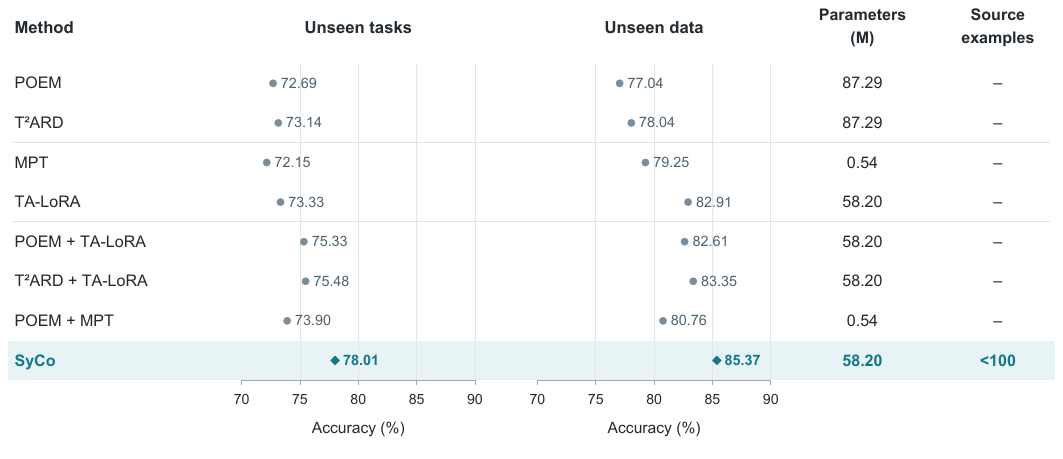}
\caption{Benchmark accuracy and adaptation resources. Mean accuracy on unseen tasks and perturbed held-out inputs from source tasks is shown for each method. Circles denote comparison methods and diamonds denote SyCo. Numerical labels give the mean accuracy in each setting, with both plots using the same horizontal scale. The two right-hand columns report parameter counts and source-example counts. SyCo uses a fixed buffer of fewer than 100 labelled source examples during adaptation.}
\label{fig:budget_comparison}
\end{figure}

SyCo achieves the highest mean accuracy among the benchmark methods in both evaluations with 58.20M inserted adapter parameters, comparable to hybrids sharing its TA-LoRA initialization (Fig.~\ref{fig:budget_comparison}). Learnable coefficients add less than 0.01\% to the adapter parameter count. Its source buffer contains fewer than 100 examples selected offline and reused throughout adaptation, so storage does not grow with the target stream. Together with the guardrail ablation (Fig.~\ref{fig:LSC3}c,d), these results support selective updates with limited source references for target adaptation.

\FloatBarrier

\section{Discussion}

Our results show that target adaptation benefits from coordinating selective parameter updates with evidence-guided regulation of update strength. This coordination allows source-derived structure to be selectively reused and modified while regulating the influence of uncertain learning signals. Across the MOA benchmark, the full framework achieved mean accuracies of 78.01\% on unseen tasks and 85.37\% on perturbed held-out inputs from source tasks, outperforming the strongest evaluated hybrid baselines. The hybrid comparisons use the same TA-LoRA initialization method, supporting the contribution of the subsequent adaptation procedure beyond initialization alone.

The ablations clarify the contributions of the two controls. Removing the Rac1-inspired confiner reduced target accuracy, and matched-ratio comparisons showed that the choice of plastic components matters beyond the amount of adaptable capacity. At the same update ratio, the proposed mask achieved higher mean accuracy on unseen tasks than IPC (78.01\% versus 77.30\%), whereas IPC achieved higher mean accuracy on unseen data (85.65\% versus 85.37\%). These comparisons vary the masking rule within the same framework. They support the utility of selective plasticity without establishing the proposed rule as universally optimal or coefficient magnitude as a universal measure of knowledge importance. Separately, replacing MAPK regulation with a fixed learning rate or removing individual evidence-processing components reduced performance. Under confidently incorrect predictions, the lower mean learning-rate multiplier (0.24 versus 0.57 for confidence-only control) indicates more conservative updating in the evaluated conditions. Together with the objective ablations, these findings support complementary contributions from parameter selection, update regulation and the learning signal.

The biological motivation provides a rationale for coordinating parameter selection with evidence-guided regulation of update strength. Rac1-related forgetting and memory-trace redistribution motivate selective retention and modification of model components \cite{zhongyi-elife,rac1-active-foregeting1,rac1-space}. MAPK-related signal processing motivates integrating multiple inputs over time and regulating the strength of the resulting response \cite{mapk-filter,mapk-gate2,mapk-amplify1,mapk-amplify2,mapk-integrate,mapk-gate1}. SyCo translates these ideas into constraints on adaptable components and regulation of the learning rate. Restricting the update location does not prevent misleading predictions from driving changes, while evidence about prediction reliability does not specify which source-derived components should be changed. Coordinating these controls provides a computational interpretation of the biological inspiration. The molecular observations motivate the design, and the empirical comparisons assess its value for target adaptation. The controller regulates updates using reliability-related evidence, but does not guarantee that every permitted update is beneficial.

Source knowledge retention in this work serves target adaptation rather than a separate objective of maintaining all previous capabilities. The stored source learners remain available separately, and the evaluation measures target-task performance. The current results cover 18 NLP datasets, including perturbed held-out source-task inputs, and depend on source retrieval and target-task descriptions. SyCo uses a small, fixed buffer of labelled source examples during adaptation, and the guardrail ablation shows that this source supervision contributes to target performance. The reported parameter and buffer counts characterize resource use, but do not establish a runtime or computational-efficiency advantage. Within this target-adaptation setting, useful next steps include examining source retrieval under larger source collections, adaptation to naturally occurring input changes and transfer to other model families. The present results support coordinated control of update location and strength as a practical approach to improving target adaptation under uncertain learning signals.

\section{Methods}

\subsection{Problem Formulation}
\label{sec:moa_formulation}

We study the MOA problem, a paradigm that integrates the objectives of MTL and TTA to model realistic deployment scenarios. 
TTA aims to improve the generalizability of a pretrained model when it is deployed on inputs whose distribution differs from that of the training data, using only unlabeled examples.
Let the labeled source dataset be $\mathcal{S} = \{(\boldsymbol{x}_i, \boldsymbol{y}_i)\}_{i=1}^M$ drawn from a distribution $p_{\text{src}}(\boldsymbol{x}, \boldsymbol{y})$.
At deployment time, the model is evaluated on a test stream $\mathcal{T}=\{\boldsymbol{x}_j\}_{j=1}^{N}$ sampled from a target distribution $p_{\text{tgt}}(\boldsymbol{x})$ differing from $p_{\text{src}}(\boldsymbol{x})$, where ground-truth labels are unavailable.
For LLMs, a common TTA paradigm partitions the model parameters into frozen backbone parameters $\boldsymbol{\theta}_f$ and a small set of adaptable parameters $\boldsymbol{\theta}_a$ \cite{houlsby2019icml,liu2022acl}.
The backbone $\boldsymbol{\theta}_f$ is kept fixed after pre-training, while $\boldsymbol{\theta}_a$ is initialized before deployment and then updated online.
We denote the model prediction on input $\boldsymbol{x}$ under parameters $(\boldsymbol{\theta}_f,\boldsymbol{\theta}_a)$ by $f(\boldsymbol{x};\boldsymbol{\theta}_f,\boldsymbol{\theta}_a)$.
Given the target test set $\mathcal{T}$, TTA updates the adaptable parameters $\boldsymbol{\theta}_a$ by minimizing an unsupervised objective over the target inputs, which can be formulated as follows:

\begin{equation}
\begin{aligned}
& \min_{\boldsymbol{\theta}_a} \,
  \mathcal{L}_{\text{TTA}}(\boldsymbol{\theta}_a; \mathcal{T}), \\
& \text{with}\quad
  \mathcal{L}_{\text{TTA}}(\boldsymbol{\theta}_a; \mathcal{T}) = \frac{1}{N} \sum_{j=1}^{N}
    \ell\big(f({\boldsymbol{x}}_j; \boldsymbol{\theta}_f, \boldsymbol{\theta}_a)\big),
\end{aligned}
\label{eq:tta_obj}
\end{equation}
where the loss $\ell(\cdot)$ is defined solely in terms of the model predictions and does not use ground-truth labels. 
The MOA setting generalizes TTA to a more realistic and challenging scenario involving multiple source tasks and open-set intents. It consists of two sequential phases:  (1) Multi-task Training: The model is first trained on \(K\) labeled source tasks \(\mathcal{S} = \{\mathcal{S}^{(k)}\}_{k=1}^K\), where each task dataset \(\mathcal{S}^{(k)} = \{(\boldsymbol{x}_i^{(k)}, \boldsymbol{y}_i^{(k)})\}_{i=1}^{N_k}\) is drawn from a distribution \(p_{\text{src}}^{(k)}(\boldsymbol{x}, \boldsymbol{y})\). This yields an initialized parameter set including frozen backbone parameters $\boldsymbol{\theta}_f$, shared trainable parameters $\boldsymbol{\theta}^{(0)}_a$ updated across all tasks, and task-specific parameters $\boldsymbol{\theta}^{(k)}_a$ for each task $k$, which are trained following Eq. (\ref{eq:mtl_obj}). 
\begin{equation}
\begin{aligned}
& \min_{\boldsymbol{\theta}^{(0)}_a,\, \{\boldsymbol{\theta}^{(k)}_a\}_{k=1}^K}
  \sum_{k=1}^{K} \lambda_k \,
  \mathcal{L}^{(k)}_{\mathrm{MTL}}(\boldsymbol{\theta}^{(0)}_a, \boldsymbol{\theta}^{(k)}_a; \mathcal{S}^{(k)}), \\
& \text{with}\quad
  \mathcal{L}^{(k)}_{\mathrm{MTL}}(\boldsymbol{\theta}^{(0)}_a, \boldsymbol{\theta}^{(k)}_a; \mathcal{S}^{(k)}) 
  = \frac{1}{N_k} \sum_{i=1}^{N_k}
    \ell_k\big(f(\boldsymbol{x}_i^{(k)}; \boldsymbol{\theta}_{f}, \boldsymbol{\theta}^{(0)}_a, \boldsymbol{\theta}^{(k)}_a),\, \boldsymbol{y}_i^{(k)}\big).
\end{aligned}
\label{eq:mtl_obj}
\end{equation}
\noindent where $\lambda_k \ge 0$ controls the relative importance of task $k$, and $\ell_k(\cdot)$ denotes the task-specific loss (e.g., cross-entropy for classification). 
(2) Open-set Test-Time Adaptation: The model is then deployed on a non-stationary, unlabeled target stream $\mathcal{T}$. Critically, the data in $\mathcal{T}$ are drawn from a mixture distribution $p_{\text{tgt}}$ whose support may partially overlap with the source tasks, combine them, or include entirely novel tasks with disjoint label or intent spaces. The mapping between source and target tasks is unknown a priori and is many-to-many.
The objective of MOA is to adapt the parameters ($\boldsymbol{\theta}^{(0)}_a$, $\{\boldsymbol{\theta}^{(k)}_a\}_{k=1}^K$) online by processing the stream $\mathcal{T}$ as in standard TTA, making full use of the knowledge from multiple source tasks. The core challenge is to coordinate which parameters may change and how strongly uncertain learning signals drive those changes, so that source knowledge supports target adaptation. Each target learner is initialized from a retrieved source learner, which remains stored separately. Our proposed algorithm, SyCo, addresses this challenge through biologically inspired mechanisms for strategic update confinement and evidence-guided regulation, detailed in the following sections.

\subsection{The SyCo Framework}
Under the MOA setting introduced above, we propose SyCo, a biologically inspired learner comprising two synergistic modules. The Rac1 module facilitates stable local adaptation. Given source tasks $\mathcal{S}=\{\mathcal{S}^{(k)}\}_{k=1}^K$ and their associated task-specific learners $\boldsymbol{\theta}_{\text{src}}=\{\boldsymbol{\theta}_{\text{src}}^{(k)}\}_{k=1}^K$ learned by a shared MTL backbone, we first identify a dedicated subspace of Low-Rank Adaptation (LoRA) \cite{lora}. This subspace is selectively reset and activated to open effective signal channels for a target unseen-task $\mathcal{T}^{(m)}$. In parallel, the MAPK module regulates update strength through temporal filtering, evidence-dependent gating, and gain modulation.
After temporal filtering, partial activation requires improved consistency together with either decreased entropy or increased likelihood, but not both. Full activation requires both decreased entropy and increased likelihood, irrespective of the consistency indicator.
Activation events modulate the learning-rate multiplier according to Eq.~\eqref{eq:learning_rate}. This multiplier scales gradient updates within the low-rank components selected by the Rac1 module. 

To derive learning signals and promote consistency, we optimize a composite objective $\mathcal{L}_\text{SC3} = \lambda_1\mathcal{L}_{\text{prob}} + \lambda_2\mathcal{L}_{\text{proc}} + \lambda_3\mathcal{L}_{\text{guard}}$. The problem level consistency term $\mathcal{L}_{\text{prob}}$ stabilizes intent under perturbations, the process level term $\mathcal{L}_{\text{proc}}$ derives self-calibrated reasoning trajectories, and the source domain guardrail $\mathcal{L}_{\text{guard}}$ uses labelled source examples to discourage drift from source-derived knowledge during target adaptation. Furthermore, to facilitate the reuse of $\boldsymbol{\theta}_{\text{src}}$, we introduce a format alignment step that rewrites the target prompt into a canonical \texttt{system}--\texttt{user} structure. Specifically, we first retrieve the most similar source task by embedding similarity, then map the target request into the retrieved task's canonical template: the \texttt{system} message specifies the role, output constraints, and decision protocol, while the \texttt{user} message carries the instance-specific input in a standardized slot. 

The whole pipeline of SyCo is shown in Fig.~\ref{fig:framework}.

\subsection{Rac1-inspired Subspace Reconfiguration}
In SyCo, the Rac1 module resets the coefficients of selected low-rank components and opens them for target adaptation, as shown in Fig. \ref{fig:rac1}. For a given source task and a given layer, the layer weight is decomposed into a shared frozen pre-trained weight matrix $\boldsymbol{W}^{(0)}$ and a task-specific low-rank update $\Delta \boldsymbol{W}$ learned via LoRA:
\begin{equation}
   \boldsymbol{W} = \boldsymbol{W}^{(0)} + \Delta \boldsymbol{W} = \boldsymbol{W}^{(0)} + \boldsymbol{B} \boldsymbol{A}^\top,
\end{equation}
where $\boldsymbol{A} \in \mathbb{R}^{d_{\text{in}} \times r}$ and $\boldsymbol{B} \in \mathbb{R}^{d_{\text{out}} \times r}$ are rank-$r$ low-rank parameters, with $d_{\text{in}}$ and $d_{\text{out}}$ denoting the input and output dimensions of $\boldsymbol{W}^{(0)}$. To enable the Rac1 module, we reparameterize the source-task low-rank update at this layer using a low-rank parameterization with learnable coefficients for each task:

\begin{equation}
    \boldsymbol{W} = \boldsymbol{W}^{(0)} + \Delta \boldsymbol{W} = \boldsymbol{W}^{(0)} + \boldsymbol{U} \boldsymbol{\Sigma} \boldsymbol{V}^{\top},
\end{equation}

\noindent  where $\boldsymbol{U} \in \mathbb{R}^{d_\text{out} \times r}$ and $\boldsymbol{V} \in \mathbb{R}^{d_\text{in} \times r}$ are low-rank matrices, and $\boldsymbol{\Sigma} = \text{diag}(\sigma_1, \dots, \sigma_r)$ is a diagonal matrix of learnable coefficients. The diagonal is represented by a learnable coefficient vector $\boldsymbol{\sigma} = (\sigma_1, \dots, \sigma_r)$, and the low-rank factors are optimized directly without computing an explicit SVD of a dense update matrix. This SVD-inspired parameterization imposes no orthogonality constraints on the factors, and its learned coefficients need not be singular values. At the start of source training, $\boldsymbol{U}$ and $\boldsymbol{V}$ are initialized with Gaussian noise and $\boldsymbol{\Sigma}$ is set to zero, so that the LoRA branch initially contributes nothing to the model output ($\Delta \boldsymbol{W} = \boldsymbol{0}$). Target adaptation subsequently starts from the trained source adapter, as described below.

In a standard SVD, each singular component contributes energy equal to its squared singular value, so retaining the leading components minimizes Frobenius-norm reconstruction error at a fixed rank. This provides a heuristic motivation for Rac1\textquotesingle s coefficient-based protection rule, whose utility in the learned parameterization is evaluated empirically. Coefficient magnitude is parameterization-dependent and is not an invariant measure of component energy or source-task importance. By sorting components in descending order of $|\sigma_j|$ in $\boldsymbol{\sigma}$, we select the top $r^* = \lfloor (1 - \alpha) r \rfloor$ directions ($\mathcal{I}_{\mathrm{keep}}$) for protection during source-to-target adaptation. Here, $\alpha \in [0,1]$ explicitly specifies the activated gradient ratio in the low-rank update. Tail components remain plastic, and the matched-ratio ablation supports this selection rule in our learned parameterization (Fig.~\ref{fig:rac1}e). 

When adapting to a novel task, SyCo employs an encoder to obtain the task embedding. It then retrieves the source adapter $\boldsymbol{\theta}_{\mathrm{src}}^{(k^*)}$ whose embedding is most similar to the task embedding (e.g., under cosine similarity), and initializes the target adapter as $\boldsymbol{\theta}_{\mathrm{tgt}} \leftarrow \boldsymbol{\theta}_{\mathrm{src}}^{(k^*)}$. Source learners remain stored separately and unchanged. Crucially, we treat the tail directions ($j \notin \mathcal{I}_{\mathrm{keep}}$) as plastic slots by resetting their coefficients to zero, thereby clearing signal channels for new gradient information. During adaptation, a gradient mask $\boldsymbol{m}$ is applied to protect the core directions ($m_j=0$ for $j \in \mathcal{I}_{\mathrm{keep}}$) while allowing the plastic slots ($m_j=1$ for $j \notin \mathcal{I}_{\mathrm{keep}}$) to be repurposed for the target task. The protected component parameters remain fixed. 

For a given target task, let $\boldsymbol{\theta}_{\mathrm{tgt},t} = \{\boldsymbol{U}_t, \boldsymbol{\Sigma}_t, \boldsymbol{V}_t\}$ be the adapter parameters at step $t$. To confine target adaptation to the selected components, we apply the mask $\boldsymbol{m}$ during the backward pass to restrict updates to the plastic directions ($j \notin \mathcal{I}_{\mathrm{keep}}$). Specifically, the update for the learnable coefficient vector $\boldsymbol{\sigma}_t$ is given by
\begin{equation} 
  \Delta \boldsymbol{\sigma}_t = -\eta_t \left[ \boldsymbol{m} \odot \operatorname{diag}\left( \boldsymbol{U}_t^\top \frac{\partial \mathcal{L}_t}{\partial \Delta \boldsymbol{W}_t} \boldsymbol{V}_t \right) \right],  \label{eq:delta_update}
\end{equation}
where $\Delta \boldsymbol{W}_t = \boldsymbol{U}_t \boldsymbol{\Sigma}_t \boldsymbol{V}_t^\top$. Similar masking is applied to the gradients of $\boldsymbol{U}_t$ and $\boldsymbol{V}_t$, ensuring that directions indexed by $\mathcal{I}_{\mathrm{keep}}$ remain fixed at their anchor values. This design keeps the protected component parameters fixed while allowing the selected components to change during target adaptation. The gradient mask restricts parameter updates without removing the protected components from the forward computation. This fixes selected parameter coordinates, but does not enforce orthogonality between plastic and protected components or guarantee preservation of source-task behavior.

\begin{theorem}[Projected stationarity under the Rac1 pathway]
\label{theorem:convergence_nonconvex}
Consider adaptation to a target task $\mathcal{T}^{(m)}$ from the source-derived initialization $\boldsymbol{\theta}_m^{(0)}$. Let the fixed target objective $\mathcal{L}_m$ be $\beta$-smooth and bounded below by $L_{\inf}$, and let $P_m$ be the orthogonal projector onto the plastic parameter subspace selected by Rac1, held fixed during the adaptation episode. For a constant step size $0<\eta\leq 1/\beta$, consider
\begin{equation}
\label{eq:rac1_stochastic_update}
\displaystyle \boldsymbol{\theta}_m^{(t+1)}=\boldsymbol{\theta}_m^{(t)}-\eta P_m g_t.
\end{equation}
Let $\mathcal{F}_t$ denote the history preceding the gradient sample $g_t$. Assume $\mathbb{E}[g_t\mid\mathcal{F}_t]=\nabla\mathcal{L}_m(\boldsymbol{\theta}_m^{(t)})$ and
\[
\displaystyle \mathbb{E}\!\left[\left\|P_m\!\left(g_t-\nabla\mathcal{L}_m(\boldsymbol{\theta}_m^{(t)})\right)\right\|^2\middle|\mathcal{F}_t\right]\leq\nu_P^2.
\]
Then, for any $N\geq 1$, the iterates satisfy
\begin{equation}
\label{eq:stationarity_rate}
\displaystyle \begin{aligned}
&\frac{1}{N}\sum_{t=0}^{N-1}\mathbb{E}\left\|P_m\nabla\mathcal{L}_m(\boldsymbol{\theta}_m^{(t)})\right\|^2\\
&\qquad\leq\frac{2[\mathcal{L}_m(\boldsymbol{\theta}_m^{(0)})-L_{\inf}]}{\eta N}+\beta\eta\nu_P^2.
\end{aligned}
\end{equation}
In particular, choosing $\eta=1/\beta$ gives an upper bound of $2\beta[\mathcal{L}_m(\boldsymbol{\theta}_m^{(0)})-L_{\inf}]/N+\nu_P^2$.
\end{theorem}
Theorem~\ref{theorem:convergence_nonconvex} provides an optimization bound for updates within the selected adaptation subspace under the stated assumptions. The bound gives an $O(1/N)$ initial-gap term and a fixed-step noise residual, recovering projected stationarity in the noise-free case (Supplementary Note~2).

\subsection{MAPK-inspired Reliability-Aware Modulation}
Complementing the Rac1-allocated subspace, the MAPK module acts as a tiered controller to regulate update strength based on evidence signals. The controller temporally filters prediction-based evidence, applies a combination-dependent activation rule and scales the learning rate accordingly, as shown in Fig. \ref{fig:MAPK}.

To implement aggregated gating for reliability-aware adaptation, we define three evidence signals, i.e., entropy $\mathcal{H}_t$, likelihood $\mathcal{P}_t$, and consistency score $\mathcal{C}_t$, and denote their relative improvements as 
$r_{\mathcal{H}}(t) = \mathcal{H}_{t-1} - \mathcal{H}_t$, 
$r_{\mathcal{P}}(t) = \mathcal{P}_{t} - \mathcal{P}_{t-1}$, and 
$r_{\mathcal{C}}(t) = \mathcal{C}_t - \mathcal{C}_{t-1}$.
A positive $r_s(t)$ for $s \in \{\mathcal{H}, \mathcal{P}, \mathcal{C}\}$ indicates a favorable evidence transition, providing evidence for learning-rate modulation.
Let $E_{s}(t) = \mathbb{I}(r_{s}(t) > 0)$ be the corresponding activation indicators, which indicate whether each signal provides positive evidence at step $t$.
Building on these indicators, we define two tiered activation events to regulate the intensity of adaptation:
\begin{align}
    A_p(t) &= \mathbb{I}\left\{\bigl[E_{\mathcal{H}}(t) \oplus E_{\mathcal{P}}(t)\bigr] \wedge E_{\mathcal{C}}(t)\right\} \label{eq:partial} \\*
    A_f(t) &= \mathbb{I}\{ E_{\mathcal{H}}(t) \wedge E_{\mathcal{P}}(t) \}. \label{eq:full}
\end{align}

Eq.~(\ref{eq:partial})--(\ref{eq:full}) implement a signal-hierarchy gating rule motivated by an information-theoretic perspective that distinguishes between distribution-intrinsic uncertainty signals and cross-generation statistical cues.
$E_{\mathcal{H}}$ and $E_{\mathcal{P}}$, are first-order indicators derived from a single predictive distribution and thus provide immediate, high-frequency evidence about the model's instantaneous uncertainty.
In contrast, $E_{\mathcal{C}}$ is a higher-order cue obtained from multiple stochastic generations, reflecting a stable self-consistent regime rather than the instantaneous reliability of the update direction.

Accordingly, the partial-update switch $A_p(t)$ in Eq.~(\ref{eq:partial}) is activated when the higher-order consistency cue $E_{\mathcal{C}}(t)$ is present, but the core first-order indicators provide only partial distribution-intrinsic evidence (i.e., either $E_{\mathcal{H}}(t)$ or $E_{\mathcal{P}}(t)$ is satisfied, but not both), yielding the condition $[E_{\mathcal{H}}(t) \oplus E_{\mathcal{P}}(t)] \wedge E_{\mathcal{C}}(t)$. The full-update switch $A_f(t)$ in Eq.~(\ref{eq:full}) is activated when both first-order indicators are concurrently satisfied, i.e., $E_{\mathcal{H}}(t) \wedge E_{\mathcal{P}}(t)$.

To implement low-pass filtering, we apply a persistence-based temporal filter with window length $l$ and persistence ratio $\kappa \in (0,1]$.
A signal is activated only when it is positive at $\lceil\kappa l\rceil$ or more positions in the window, thereby suppressing transient fluctuations.
The corresponding smoothed indicator is defined as
\begin{equation}
    \widetilde{E}_s(t)
    = \mathbb{I}\Bigl\{
        \sum_{j=0}^{l-1} E_s(t-j) \ge \lceil \kappa l \rceil
      \Bigr\},
    \quad s \in \{\mathcal{H}, \mathcal{P}, \mathcal{C}\}.
    \label{eq:sumE}
\end{equation}

We apply the logical rules from Eq.~\eqref{eq:partial}--\eqref{eq:full} to the smoothed indicators $\widetilde{E}_s(t)$ to obtain the windowed activation events $\widetilde{A}_p(t)$ and $\widetilde{A}_f(t)$. Without activation, the baseline learning rate applies. These windowed events ensure that the MAPK module responds to persistent trends rather than instantaneous fluctuations. To maintain temporal consistency, we couple these activations with gradients over the same horizon $l$, aligning the effective gradient with the reliability evidence. The MAPK pathway modulates the learning rate as
\begin{equation}
\eta_t
= \eta_0 \bigl( \gamma_0
+ \omega_{\text{amp}} \gamma_1 \widetilde{A}_p(t)
+ \omega_{\text{amp}} \gamma_2 \widetilde{A}_f(t) \bigr),
\label{eq:learning_rate}
\end{equation}
where $\eta_0$ is the base learning rate. The scalar $\gamma_0 \in (0,1)$ provides a conservative baseline in the absence of activation, while $\gamma_1$ and $\gamma_2$ ($\gamma_2 > \gamma_1 > 0$) represent the gains for partial and full activation, respectively. The fixed amplification scale $\omega_{\text{amp}}$ increases the activation-dependent contributions to the learning-rate multiplier. The controller therefore modulates update strength, retaining a nonzero baseline when activation conditions are unmet.

\subsection{Structured Cognitive Consistency Constraint ($\mathcal{L}_\text{SC3}$)}
In non-stationary streams, conventional self-training objectives often fail to distinguish core task intent from superficial linguistic variations, causing models to overfit to noise and triggering catastrophic representation drift. To organize learning signals from target predictions, input variations and labelled source examples, we formulate $\mathcal{L}_\text{SC3}$. The overall adaptation objective is defined as
\begin{equation}
\mathcal{L}_\text{SC3} = \lambda_1 \mathcal{L}_{\text{prob}} + \lambda_2 \mathcal{L}_{\text{proc}} + \lambda_3 \mathcal{L}_{\text{guard}},
\label{eq:loss_total}
\end{equation}
where $\lambda_1, \lambda_2, \lambda_3$ are weighting coefficients. This tripartite loss decomposes the adaptation objective into three synergistic components: problem understanding loss $\mathcal{L}_{\text{prob}}$, process-based reasoning loss $\mathcal{L}_{\text{proc}}$, and source-domain guardrail $\mathcal{L}_{\text{guard}}$, as is shown in Fig. \ref{fig:LSC3}. While the Rac1 module constrains updates to a low-rank subspace, the $\mathcal{L}_{\text{guard}}$ term provides a source-data constraint intended to limit drift from source-derived knowledge during target adaptation.

The first component, $\mathcal{L}_{\text{prob}}$, promotes paraphrase invariance by ensuring that semantically equivalent prompts yield consistent behaviors. It comprises two complementary terms:
\begin{equation}
\begin{aligned}
\mathcal{L}_{\text{prob}} =
&\underbrace{\mathrm{CE}\bigl(p_{\boldsymbol{\theta}}(\cdot \mid \tilde{q}),\, q\bigr)}_{
  \text{Paraphrase Supervision}
}
+  \underbrace{\mathrm{InfoNCE}\bigl(h(q),\, h(\tilde{q}),\, h(q^{-})\bigr)}_{
  \text{Semantic Contrastive Alignment}
},
\end{aligned} \label{eq:loss_question}
\end{equation}
where $q$ denotes the canonical query, $\tilde{q}$ is its paraphrased variant, and $q^{-}$ is a negative sample drawn from the current batch.
$p_{\boldsymbol{\theta}}(\cdot \mid \cdot)$ is the conditional next-token distribution of a decoder-only model parameterized by $\boldsymbol{\theta}$. $\mathrm{CE}(\cdot,\cdot)$ denotes the token-level cross-entropy loss between the predicted distribution and the target sequence.

The InfoNCE objective pulls the representations $h(q)$ and $h(\tilde{q})$ together in the embedding space while pushing them away from the unrelated negative $h({q}^{-})$. $h(\cdot)$ denotes the semantic representation extracted from the decoder-only model's last-layer hidden states. By combining discrete token prediction with continuous representation alignment, $\mathcal{L}_{\text{prob}}$ enforces robust problem understanding across diverse linguistic expressions.

The second component, $\mathcal{L}_{\text{proc}}$, shown in Eq.~\eqref{eq:loss_answer}, stabilizes adaptation under uncertainty by coupling pseudo-labeling with entropy-based regularization. This mechanism is specifically designed to mitigate confirmation bias---a common failure mode where models overfit to their own erroneous predictions during adaptation.
\begin{equation}
\mathcal{L}_{\mathrm{proc}}
=
\underbrace{\mathrm{CE}\!\left(p_{\boldsymbol{\theta}}(\cdot \mid q;\tau),\, a^{*}\right)}_{\text{Pseudo-Labeling}}
\;-
\underbrace{\mathcal{H}\!\left(p_{\boldsymbol{\theta}}(\cdot \mid q)\right)}_{\text{Confidence}} \label{eq:loss_answer}
\end{equation}

Here, $a^*$ is the pseudo-label derived from the model's most confident prediction. To stabilize adaptation, the negative entropy term $-\mathcal{H}(\cdot)$ acts as an anti-collapse regularizer. Specifically, $\mathcal{H}(\cdot)$ denotes the Shannon entropy, and its maximization encourages output dispersion. By penalizing overly peaky distributions, it counteracts the tendency of pseudo-labeling to produce degenerate, overconfident outputs. Together with a temperature $\tau > 1$ that softens the loss, these mechanisms aim to limit overconfident self-reinforcement during adaptation.

The final component, $\mathcal{L}_{\text{guard}}$, acts as a source-domain anchor to limit drift from source-derived knowledge during target adaptation. Target adaptation uses no target labels, but retains access to a fixed labelled source subset for the guardrail. The subset $\mathcal{S}^*$ is selected once offline using the source learners' prediction scores. We favor examples with a large score gap between the top two candidates to anchor adaptation to confident source predictions. During adaptation, source mini-batches are drawn from this fixed subset. We optimize:
\begin{equation}
\mathcal{L}_{\text{guard}}
= \mathbb{E}_{(q, a) \sim \mathcal{S}^{*}}
\, \mathrm{CE}\!\left( p_{\boldsymbol{\theta}}(\cdot \mid q),\, a \right),
\label{eq:loss_sample}
\end{equation}
where the inclusion of ground-truth source labels, i.e., $a$, provides a stable gradient signal. This mechanism provides source supervision during target adaptation and is intended to limit instability from self-training.

For notational clarity, we omit constant balancing coefficients and hyperparameters from the individual loss terms, as they are fixed across all experiments and do not affect the form of the gradients.
In our implementation, the MAPK activation events serve only to modulate the learning rate $\eta_t$ and are not backpropagated through, i.e., they do not enter the gradient of Eq.~\eqref{eq:loss_total}.

\begin{remark}[Smoothness assumption in Theorem~\ref{theorem:convergence_nonconvex}]
Theorem~\ref{theorem:convergence_nonconvex} assumes that the objective $\mathcal{L}_{\mathrm{SC3}}$ is $\beta$-smooth with respect to the parameters being optimized. A sufficient condition is that this objective is twice continuously differentiable and its Hessian is uniformly bounded on a convex neighbourhood containing the iterates.
\label{remark:smoothness}
\end{remark}

Algorithm~\ref{alg:syco} summarizes the overall SyCo procedure.

\begin{algorithm}[ht]
\caption{SyCo in the MOA Setting}
\label{alg:syco}
\SetAlgoNlRelativeSize{-1}
\SetKwInOut{Input}{Input}
\SetKwInOut{Output}{Output}

\BlankLine
Pretrain multi-source adapters $\{\boldsymbol{\theta}_{\mathrm{src}}^{(k)}\}_{k=1}^{K}$ on source tasks $\mathcal{S}$\;
Select fixed labelled subset $\mathcal{S}^{*}\subset\mathcal{S}$ once offline, favoring large top-two prediction-score gaps under the source learners\;
\textbf{Rac1 module: Structure Initialization (once)}\\
Initialize $\boldsymbol{\theta}_{\mathrm{tgt}}=\{\boldsymbol{U},\boldsymbol{\Sigma},\boldsymbol{V}\}$ by $\boldsymbol{\theta}_{\mathrm{src}}^{(k^*)}$ selected via task embedding similarity\;
For each low-rank adapter of rank $r$, sort the components by descending $|\sigma_j|$\;
$\mathcal{I}_{\mathrm{keep}} \gets$ indices of the first $\lfloor(1-\alpha)r\rfloor$ components\;
Reset $\sigma_j\gets 0$ for $j\notin\mathcal{I}_{\mathrm{keep}}$\;
$\boldsymbol{m} \gets$ gradient mask with $m_j = 0$ if $j \in \mathcal{I}_{\mathrm{keep}}$ else $1$\;

\BlankLine
\textbf{MAPK module: Test-Time Adaptation Loop}\\
\For{$t = 1$ \textbf{to} $T$}{
    Sample target mini-batch $\mathcal{B}_t$ from $\mathcal{T}$\;
    Sample source mini-batch $\mathcal{B}^{*}_t$ from the fixed subset $\mathcal{S}^{*}$\;

    \For {each $x_b$ in $\mathcal{B}_t$}{
        Convert $x_b$ to normalized query $q_b$\;
        Generate perturbed $\tilde{q}_b$ and batch-sampled negative $\tilde{q}_b^{-}$\;
        Generate $M$ candidates $a_{b,1:M}$\;
        Select pseudo-label $a_b^{*}$ by confidence score\;
    }
    $\hat{y}_t \gets \{a_b^{*}\}_{b=1}^{|\mathcal{B}_t|}$\;

    Validate $\widetilde{E}_s(t)$ for $s\in\{\mathcal{H},\mathcal{P},\mathcal{C}\}$ using Eq.~\eqref{eq:sumE}\;
    $\widetilde{A}_p(t) \gets \mathbb{I}\{[\widetilde{E}_{\mathcal{H}}(t) \oplus \widetilde{E}_{\mathcal{P}}(t)] \wedge \widetilde{E}_{\mathcal{C}}(t)\}$\;
    $\widetilde{A}_f(t) \gets \mathbb{I}\{\widetilde{E}_{\mathcal{H}}(t) \wedge \widetilde{E}_{\mathcal{P}}(t)\}$\;

    Compute $\mathcal{L}_{\text{prob}}(q, \tilde{q}, \tilde{q}^-)$ using Eq.~\eqref{eq:loss_question}\;
    Compute $\mathcal{L}_{\text{proc}}(q, a^*)$ using Eq.~\eqref{eq:loss_answer}\;
    Compute $\mathcal{L}_{\text{guard}}(\mathcal{B}^{*}_t)$ using Eq.~\eqref{eq:loss_sample}\;
    $\mathcal{L}_\text{SC3}^{(t)} \gets \lambda_1 \mathcal{L}_{\text{prob}} + \lambda_2 \mathcal{L}_{\text{proc}} + \lambda_3 \mathcal{L}_{\text{guard}}$\;

    $\eta_t \gets \eta_0\bigl(\gamma_0+\omega_{\text{amp}} \gamma_1\widetilde{A}_p(t)+\omega_{\text{amp}} \gamma_2\widetilde{A}_f(t)\bigr)$\;

    \For{\text{each} $\boldsymbol{c}\,\,in\,\, \{\boldsymbol{U}, \boldsymbol{V}, \boldsymbol{\Sigma}\}$}{
        $\nabla_{\boldsymbol{c}} \mathcal{L}_\text{SC3}^{(t)} \gets \nabla_{\boldsymbol{c}} \mathcal{L}_\text{SC3}^{(t)} \odot \boldsymbol{m}$\;
        $\boldsymbol{c}^{(t+1)} \gets \boldsymbol{c}^{(t)} - \eta_t \nabla_{\boldsymbol{c}} \mathcal{L}_\text{SC3}^{(t)}$\;
    }
}
\Return $\{\hat{y}_t\}_{t=1}^{T}$
\end{algorithm}

\subsection{Experimental Setup}
\subsubsection{Datasets and Protocols} To simulate the MOA challenges faced by LLMs in real deployment, we utilize 18 NLP datasets~\cite{dataset, talora} under two protocols:

\textbf{(i) Unseen-task (cross-task generalization).} We fine-tuned the backbone on eight source tasks and adapted the resulting source-informed learners to ten unseen target tasks: ChnSent, DRCD, TNews, COTE-BD, OCNLI, FinRE, LCQMC, CCPM, CoteDp and C3. This protocol evaluates adaptation to target tasks absent from source training under the MOA setting.

\textbf{(ii) Unseen-data (distributional robustness).} We apply additional perturbations to held-out inputs from the same eight source tasks. This protocol evaluates target adaptation to the resulting input shifts within source-task categories.

Detailed dataset statistics and training/test splits are provided in Supplementary Information, Supplementary Note~3.

\subsubsection{Baselines}
We compare SyCo against three categories of baselines: (i) TTA-only methods, (ii) MTL-only models, and (iii) MTL combined with TTA.

\textbf{(i) TTA-only baselines.} We include representative methods such as TENT \cite{tent}, ClusT3 \cite{clust3}, FOA \cite{foa}, POEM \cite{poem}, and T$^2$ARD \cite{t2ard}, which adapt LLMs via single-task supervision followed by test-time updates without leveraging multi-source knowledge.

\textbf{(ii) MTL-only baselines.} We consider Prefix Tuning (PT) \cite{pt} methods including SPoT \cite{spot}, ATTEMPT \cite{attempt}, MPT \cite{mpt}, and TA-LoRA \cite{talora}, which are trained via cross-task shared regularization and tested in a zero-shot manner. We prioritize PT baselines as their trainable prefixes inherently encode task-specific knowledge, naturally facilitating the decoupling of shared and task-specific representations in multi-task learning.

\textbf{(iii) MTL+TTA baselines.} We compare SyCo with hybrid methods that combine the same TA-LoRA initialization method with the TTA objectives from (i). These combinations serve as competitive baselines by bridging multi-task knowledge with online adaptation, effectively covering the MOA landscape where specialized methods are currently lacking.

\subsubsection{Implementation Details}
For both evaluation protocols, the guardrail used a fixed buffer of fewer than 100 labelled source-training examples, selected once before target adaptation by favoring large top-two prediction-score gaps under the source learners. Each adaptation step sampled a mini-batch from this buffer without rescoring or reselecting examples from the full source datasets.

\begin{table}[h]
\centering
\caption{Implementation hyperparameters for Rac1/MAPK-inspired modules.}
\label{tab:impl_details}
\small
\renewcommand{\arraystretch}{1.15}
\setlength{\tabcolsep}{6pt}
\begin{tabular}{@{}
>{\centering\arraybackslash}p{0.18\columnwidth}
>{\centering\arraybackslash}p{0.40\columnwidth}
>{\centering\arraybackslash}p{0.34\columnwidth}
@{}}
\toprule
\textbf{Module} & \textbf{Hyperparameter} & \textbf{Value} \\
\midrule
Rac1 & mask ratio $\alpha$ & $0.1$ \\
\midrule
MAPK & base learning rate $\eta_0$ & $5\times10^{-4}$ \\
MAPK & gains $(\gamma_0,\gamma_1,\gamma_2)$ & $(0.1,\,0.5,\,1.0)$ \\
MAPK & smoothing window $l$ & 50 \\
MAPK & threshold $\kappa$ & $0.8$ \\
MAPK & amplification scale $\omega_{\text{amp}}$ & $1.2$ \\
\midrule
Loss & weights $(\lambda_1,\lambda_2,\lambda_3)$ & $(0.2,\,0.7,\,0.1)$ \\
Reg. & entropy temperature $\tau$ & $\tau=1.2$ \\
\bottomrule
\end{tabular}
\end{table}

We summarize the key hyperparameters for SyCo in Table~\ref{tab:impl_details}. For Rac1, a light masking strategy with a low mask ratio $\alpha$ is applied to preserve structural integrity. In MAPK, we establish a base learning rate $\eta_0$ and employ temporal smoothing governed by a window length $l$ and a persistence threshold $\kappa$ to stabilize activation indicators. To modulate the step size, we apply a three-stage gain $(\gamma_0, \gamma_1, \gamma_2)$ and introduce an amplification scale $\omega_{\text{amp}}$ to adaptively enhance the intensity of active updates. The composite loss function weights $(\lambda_1, \lambda_2, \lambda_3)$ are configured to prioritize the process-level calibration task ($\mathcal{L}_{\text{proc}}$), while an entropy temperature $\tau > 1$ is adopted for regularization to prevent prediction over-confidence. All parameters are optimized via grid search on the validation set.

\section*{Acknowledgments}
	This work was supported in part by National Natural Science Foundation of China (No. U25B20273, H. M.), Beijing Natural Science Foundation (No. L257003, H. M.), National Natural Science Foundation of China (No. 62227801, H. M.), National Natural Science Foundation of China (No. 62172036, T. H.), and National Science and Technology Major Project (No. 2022ZD0116305, T. H.).

\section*{Author contributions}
X.Z. and T.H. contributed equally to this work. X.Z. and T.H. conceived the research, designed the overall framework, and wrote the manuscript. X.Z. implemented the algorithms and conducted the experiments. J.L. assisted with data collection, preprocessing, and benchmark evaluations. H.M. and Q.Z. supervised the project. All authors discussed the results and contributed to the final manuscript.

\section*{Data availability}
The textual datasets and corpora used for training and evaluating the models in this study are derived from publicly available sources. Complete descriptions, direct access links, and preprocessing specifications for all public text corpora are provided in the Supplementary Information. The source data and experimental records generating the figures and tables presented in this paper will be deposited in the Figshare repository with a persistent DOI upon formal publication of the manuscript.

\section*{Code availability}
The custom code and algorithmic implementations developed in this study are encapsulated into a Code Ocean compute capsule to enable direct execution and verification by the editors and referees during the peer review process. Upon formal acceptance and publication, the permanent Code Ocean capsule (with an assigned DOI) and the corresponding open-source repository will be made publicly available under an open-source license.

\section*{Competing Interests}
	The authors declare no competing interests.

\end{document}


\setcounter{section}{0}
\setcounter{figure}{0}
\setcounter{table}{0}
\setcounter{equation}{0}
\setcounter{page}{1} 

\renewcommand{\figurename}{Supplementary Fig.}
\renewcommand{\tablename}{Supplementary Table}

\renewcommand{\thefigure}{\arabic{figure}}
\renewcommand{\thetable}{\arabic{table}}
\renewcommand{\theequation}{S\arabic{equation}}

\begin{center}
    \Large\bfseries Supplementary Information
    \par\medskip\normalsize Biologically Inspired Coordination of Parameter Selection and Update Strength for Language Model Adaptation
\end{center}
\vspace{1em}

\section*{Supplementary Note 1: Backbone and LLM Ablation} \label{app:backnone}

Our target-adaptation framework builds on a multi-task backbone trained with TA-LoRA~\cite{talora}, a task-adaptive low-rank adapter module. TA-LoRA handles task heterogeneity by inserting modular low-rank adapters into the final $L$ layers of a frozen LLM. For a layer and task $k$, the adapted weight is
\begin{equation}
\boldsymbol{W}^{(k)} = \boldsymbol{W}^{(0)} + \Delta \boldsymbol{W}^{(k)} = \boldsymbol{W}^{(0)} + \boldsymbol{B}(\boldsymbol{u}^{(k)} \boldsymbol{v}^{(k)\top}), \label{eq:ta-lora}
\end{equation}
where $\boldsymbol{B}$ is a shared low-rank matrix capturing global knowledge, while $(\boldsymbol{u}^{(k)}, \boldsymbol{v}^{(k)})$ are task-specific rank-1 parameters modeling local variations. This decomposition reduces the number of trainable parameters and regularizes task-specific updates.

TA-LoRA further stabilizes multi-task pretraining by zero-initializing the adapter branch and using a learnable gate to gradually increase its contribution, which mitigates early-stage interference. In this work, we reuse the pretrained TA-LoRA backbone as the starting point for test-time adaptation, without modifying its pretraining protocol. All components of SyCo are built on top of this shared multi-source substrate.

We perform ablation experiments on the TA-LoRA component to validate its role as a fundamental catalyst for multi-source synergy within SyCo. This analysis answers two pivotal questions: first, whether TA-LoRA improves performance across the evaluated model families; and second, how its contribution evolves as the model's computational capacity increases.
\begin{table}[b]
\centering
\caption{Ablation of \textbf{backbones} under the MOA setting. All configurations share the same test-time adaptation module and differ only in whether the multi-source learner uses TA-LoRA. We report Unseen-data/task Avg. with and without TA-LoRA.}
\label{tab:ablate_llm}
\begin{tabular}{@{}l|c|cc|cc@{}}
\toprule
\multirow{2}{*}{\textbf{Backbone LLM}} & 
\multirow{2}{*}{\textbf{\#Params}} & 
\multicolumn{2}{c|}{\textbf{w/o TA-LoRA}} & 
\multicolumn{2}{c}{\textbf{+ TA-LoRA}} \\
\cmidrule(lr){3-4}\cmidrule(lr){5-6}
 & & Unseen-data & Unseen-task & Unseen-data & Unseen-task\\
\midrule
Qwen3         & 8B & 82.91 & 73.34 & 85.37 & 78.01 \\
LLaMA3.1      & 8B & 83.50 & 74.25 & 86.02 & 77.95 \\
DeepSeek-LLM  & 7B & 83.00 & 73.90 & 85.60 & 77.27 \\
Mistral       & 7B & 80.37 & 71.90 & 83.50 & 76.46 \\
\bottomrule
\end{tabular}
\end{table}

\begin{figure}[tbp]
    \centering
    \includegraphics[width=0.7\textwidth]{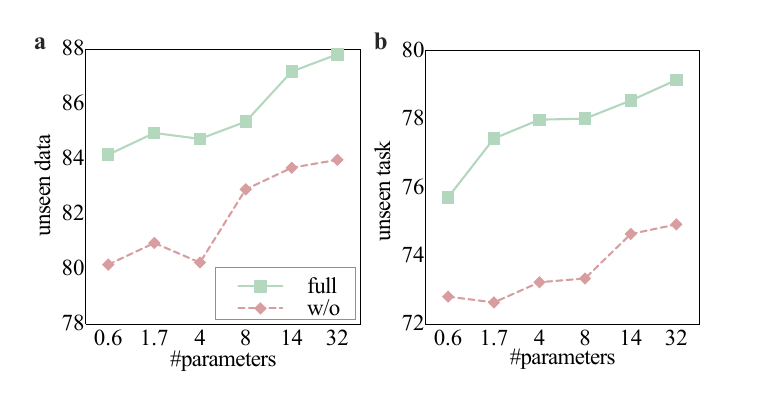}
    \caption{Scaling curves of Qwen3 models with and without TA-LoRA under the MOA setting. (a) Unseen-data performance as a function of model size. (b) Unseen-task performance as a function of model size.}
    \label{fig:qwen_scaling}
\end{figure}

Supplementary Table~\ref{tab:ablate_llm} and Supplementary Fig.~\ref{fig:qwen_scaling} are ablations for the TA-LoRA component in SyCo, answering two complementary aspects of its effectiveness under the MOA protocol and the same test-time adaptation setting.

Supplementary Table~\ref{tab:ablate_llm} tests whether TA-LoRA provides consistent SyCo gains across backbone families. We vary only the backbone LLM and compare SyCo variants trained w/o TA-LoRA to those trained with TA-LoRA on Qwen3, LLaMA3.1, DeepSeek-LLM, and Mistral. The results show that adding TA-LoRA consistently improves both unseen-task and unseen-data performance across all backbones. These comparisons support the contribution of multi-source initialization within SyCo across the evaluated backbones.
Supplementary Fig.~\ref{fig:qwen_scaling} then focuses on a single family (Qwen) to examine how TA-LoRA's contribution to SyCo changes with model size. Performance generally improves with model size, with local fluctuations. Full SyCo consistently outperforms the w/o TA-LoRA variant in both evaluations at every tested scale (Supplementary Fig.~\ref{fig:qwen_scaling}a,b), supporting the benefit of multi-source initialization across model sizes.

\section*{Supplementary Note 2: Proof of Theorem 1} \label{app:theorem_proof}
We analyze the sequence of stochastic updates within the fixed plastic subspace selected by Rac1. The mask ratio determines which parameter directions remain available for adaptation. The following proof tracks the dependence on the smoothness constant, step size and gradient noise throughout the adaptation episode.

\begin{proof}
For compactness, write $L=\mathcal{L}_m$, $\theta_t=\boldsymbol{\theta}_m^{(t)}$ and $P=P_m$. Let $\mathcal{F}_t$ contain the initialization and all randomness preceding the gradient sample $g_t$. Under the assumptions of Theorem 1, $P$ is a fixed orthogonal projector, $L$ is $\beta$-smooth and bounded below by $L_{\inf}$, and the update is $\theta_{t+1}=\theta_t-\eta P g_t$, with $0<\eta\leq 1/\beta$. Conditional unbiasedness and the projected variance bound imply
\[
\displaystyle \begin{aligned}
\mathbb{E}[P g_t\mid\mathcal{F}_t]&=P\nabla L(\theta_t),\\
\mathbb{E}[\|P g_t\|^2\mid\mathcal{F}_t]
&=\|P\nabla L(\theta_t)\|^2
+\mathbb{E}[\|P(g_t-\nabla L(\theta_t))\|^2\mid\mathcal{F}_t]\\
&\leq\|P\nabla L(\theta_t)\|^2+\nu_P^2.
\end{aligned}
\]
By the descent lemma,
\begin{equation}
\label{eq:descent_lemma}
\displaystyle L(\theta_{t+1})\leq L(\theta_t)-\eta\langle\nabla L(\theta_t),P g_t\rangle
+\frac{\beta\eta^2}{2}\|P g_t\|^2.
\end{equation}
Taking conditional expectations and using $P=P^\top=P^2$ gives
\begin{equation}
\label{eq:descent_step}
\displaystyle \begin{aligned}
\mathbb{E}[L(\theta_{t+1})\mid\mathcal{F}_t]
&\leq L(\theta_t)-\eta\left(1-\frac{\beta\eta}{2}\right)\|P\nabla L(\theta_t)\|^2
+\frac{\beta\eta^2}{2}\nu_P^2\\
&\leq L(\theta_t)-\frac{\eta}{2}\|P\nabla L(\theta_t)\|^2
+\frac{\beta\eta^2}{2}\nu_P^2,
\end{aligned}
\end{equation}
where the last inequality uses $\eta\leq 1/\beta$. Taking total expectations and summing over all $N$ updates telescopes the loss differences:
\begin{equation}
\label{eq:telescoping}
\displaystyle \begin{aligned}
\frac{\eta}{2}\sum_{t=0}^{N-1}\mathbb{E}\|P\nabla L(\theta_t)\|^2
&\leq L(\theta_0)-\mathbb{E}L(\theta_N)+\frac{N\beta\eta^2}{2}\nu_P^2\\
&\leq L(\theta_0)-L_{\inf}+\frac{N\beta\eta^2}{2}\nu_P^2.
\end{aligned}
\end{equation}
Dividing by $\eta N/2$ yields
\[
\displaystyle \frac{1}{N}\sum_{t=0}^{N-1}\mathbb{E}\|P\nabla L(\theta_t)\|^2
\leq\frac{2[L(\theta_0)-L_{\inf}]}{\eta N}+\beta\eta\nu_P^2.
\]
In particular, setting $\eta=1/\beta$ gives $2\beta[L(\theta_0)-L_{\inf}]/N+\nu_P^2$. This proves the bound for the sequence of fixed-step projected stochastic updates specified in Theorem 1, since the one-step inequality holds conditionally on the entire preceding history at every step.
\end{proof}
The initial-gap contribution decreases as $O(1/N)$, while the fixed-step residual is controlled by the projected gradient noise. In the noise-free case, the bound recovers $O(1/N)$ convergence to projected first-order stationarity. More generally, the bound quantifies how progress within the admissible subspace depends jointly on initialization, step size and stochastic variability.

\begin{table}[h]
   
   \caption{Overview of the datasets utilized for training and evaluation. These datasets span diverse tasks and domains, serving as a comprehensive benchmark for assessing the performance of SyCo.}
   \centering
   \small\setlength{\tabcolsep}{3pt}
\begin{tabular}{@{}p{0.04\textwidth}p{0.13\textwidth}p{0.23\textwidth}p{0.15\textwidth}p{0.07\textwidth}p{0.26\textwidth}@{}}
   \toprule
   \textbf{ID} & \textbf{Dataset}   & \textbf{Task Type}              & \textbf{Domain}         & \textbf{Size}  & \textbf{Source/Link} \\ \midrule
   1  & AFQMC     & Semantic Matching       & Financial       & 38K   & \href{https://aclanthology.org/2020.coling-main.419/}{Xu et al. (2020)}  \\
   2  & Amazon    & Sentiment Analysis      & Shopping Reviews & 4.1M  & Not specified  \\
   3  & THUCNews  & Text Classification     & General          & 55K   & Not specified  \\
   4  & BQ        & Semantic Matching       & Financial        & 110K  & \href{https://aclanthology.org/D18-1536/}{Chen et al. (2018)}\\
   5  & CMNLI     & Natural Language Inference & General      & 404K  & \href{https://aclanthology.org/2020.coling-main.419/}{Xu et al. (2020)} \\
   6  & CMRC-2018 & Reading Comprehension   & General          & 11.9K & \href{https://aclanthology.org/2020.coling-main.419/}{Xu et al. (2020)}  \\
   7  & SanWen    & Relation Extraction     & Literature       & 16K   & \href{https://aclanthology.org/2020.coling-main.419/}{Xu et al. (2020)}  \\
   8  & COTE-MFW  & Opinion Mining          & Shopping Reviews & 37K   & \href{https://proceedings.mlr.press/v95/li18d.html}{Li et al. (2018)}  \\
   9  & ChnSent   & Sentiment Analysis      & Shopping Reviews        & 12K   & Not specified  \\
   10 & TNews     & Text Classification     & General          & 63K   & \href{https://github.com/aceimnorstuvwxz/toutiao-text-classfication-dataset}{Github}  \\
   11 & OCNLI     & Natural Language Inference & General    & 53K   & \href{https://aclanthology.org/2020.findings-emnlp.314/}{Hu et al. (2020)}  \\
   12 & LCQMC     & Question Matching       & General          & 250K  & \href{https://aclanthology.org/C18-1166/}{Liu et al. (2018)}  \\
   13 & DRCD      & Reading Comprehension   & General          & 10K   & \href{https://arxiv.org/abs/1806.00920}{Shao et al. (2018)}  \\
   14 & C3        & Multi-choice QA         & General          & 11K   & \href{https://direct.mit.edu/tacl/article/doi/10.1162/tacl_a_00305/43546/Investigating-Prior-Knowledge-for-Challenging}{Sun et al. (2018)}  \\
   15 & COTE-BD   & Aspect-based Sentiment Analysis & History & 8K    & \href{https://proceedings.mlr.press/v95/li18d.html}{Li et al. (2018)}  \\
   16 & CoteDp   & Aspect-based Sentiment Analysis & Shopping Reviews  & 25K    & \href{https://proceedings.mlr.press/v95/li18d.html}{Li et al. (2018)}  \\
   17 & FinRE     & Relation Extraction     & Financial        & 14K   & \href{https://aclanthology.org/P19-1430/}{Li et al. (2019)}  \\  
   18 & CCPM     & Semantic Matching      & Literature        & 24K   & \href{https://arxiv.org/abs/2106.01979/}{Li et al. (2021)}  \\  
   \bottomrule
   \end{tabular} \label{tab:data}
\end{table}
  
\section*{Supplementary Note 3: Dataset Details} \label{app:datasets}

This section provides detailed information about the datasets used for training, unseen-data evaluation, and unseen-task evaluation. The datasets span diverse tasks, domains, and sizes, offering a comprehensive benchmark for assessing the performance of SyCo.

\subsection*{Training Data}
We utilize the training sets of eight source tasks for multi-task learning: AFQMC, Amazon, THUCNews, BQ, CMNLI, CMRC-2018, SanWen, and COTE-MFW. These datasets encompass various tasks such as semantic matching, sentiment analysis, reading comprehension, and text classification. The diversity of these tasks enables the model to learn shared representations across tasks while capturing task-specific nuances. Additionally, small-scale datasets are augmented using techniques like up-sampling and data enhancement methods, including synonym substitution and random addition or deletion. For large-scale datasets, down-sampling is applied to achieve balance.

\textbf{-\! Task types:} These tasks are designed to evaluate different aspects of language understanding, such as recognizing semantic equivalence (e.g., AFQMC, BQ), opinion mining (e.g., COTE-MFW), and information extraction (e.g., CMRC-2018).

\textbf{-\! Domains:} The datasets cover a wide range of domains, including financial text, shopping reviews, and general literature, ensuring that the model is exposed to diverse linguistic styles and vocabularies.

\textbf{-\! Dataset statistics:} As shown in Supplementary Table~\ref{tab:data}, the sizes of the training sets vary significantly, ranging from 11.9K (CMRC-2018) to 4.1M (Amazon). This variability provides an opportunity to test the model's robustness across tasks of different scales.

\subsection*{Target Tasks of Unseen-Data}
To evaluate the model's ability to generalize to unseen-data within the same tasks, we use the validation sets of the eight source tasks (AFQMC, Amazon, THUCNews, BQ, CMNLI, CMRC-2018, SanWen, and COTE-MFW). These validation sets are held out from source training. Additional perturbations are applied to their inputs for unseen-data evaluation.

\textbf{-\! Purpose:} This evaluation assesses target adaptation to perturbed held-out inputs from tasks included in source training.

\textbf{-\! Dataset characteristics:} These datasets consist of strictly held-out samples from the source tasks, ensuring they remain completely unseen by the model during the training phase. For example, the evaluation sets for AFQMC and COTE-MFW contain 4.3K and 1.7K unseen samples, respectively.

\subsection*{Target Tasks of Unseen-Task}
To assess the model's capability to generalize across tasks, we employ the training sets of ten downstream tasks: ChnSent, DRCD, TNews, COTE-BD, OCNLI, FinRE, LCQMC, CCPM, CoteDp, and C3. These tasks are absent from the source-training task set, although broad task types and domains can overlap with those of the source tasks.

\textbf{-\! Task types:} The unseen-task includes sentiment analysis (e.g., ChnSent), natural language inference (e.g., OCNLI), and machine reading comprehension (e.g., DRCD, C3), providing target datasets outside the source-training set.

\textbf{-\! Domains:} These datasets span financial text (e.g., FinRE), shopping reviews (e.g., ChnSent), general-domain inference (e.g., OCNLI), and general knowledge (e.g., TNews, LCQMC), allowing the evaluation of domain transfer capabilities.

\textbf{-\! Dataset statistics:} As shown in Supplementary Table~\ref{tab:data}, these datasets vary in size from 8K (COTE-BD) to 250K (LCQMC), providing a challenging benchmark for cross-task generalization.

\subsection*{Analysis and Implications}
The dataset selection in Supplementary Table~\ref{tab:data} ensures a balance between task and domain diversity, as well as variability in data size. This design serves several key purposes:

\textbf{-\! Diversity in learning:} Source tasks with different domains and linguistic structures provide a basis for learning representations that can be reused during target adaptation.

\textbf{-\! Dataset coverage:} The evaluation covers datasets of different sizes. Model-size comparisons are reported separately in Supplementary Note 1.

\textbf{-\! Cross-domain generalization:} The inclusion of tasks from distinct domains (e.g., financial, review, and general text) highlights the model's ability to adapt to unseen contexts, a crucial feature for practical applications.

Together, these datasets support evaluation of source-initialized target adaptation under the unseen-task and unseen-data protocols.

\enlargethispage{2\baselineskip}